\documentclass[sn-mathphys,Numbered]{sn-jnl}

\usepackage{graphicx}%
\usepackage{multirow}%
\usepackage{amsmath,amssymb,amsfonts}%
\usepackage{amsthm}%
\usepackage{mathrsfs}%
\usepackage[title]{appendix}%
\usepackage{xcolor}%
\usepackage{textcomp}%
\usepackage{manyfoot}%
\usepackage{booktabs}%
\usepackage{algorithm}%
\usepackage{listings}%
\usepackage{epsfig}
\usepackage{booktabs}
\usepackage{multirow}
\usepackage{subcaption}
\usepackage{algorithmic}
\usepackage{newfloat}
\usepackage{caption} 
\usepackage{natbib}
\usepackage{courier} 
\usepackage{helvet}
\usepackage{times}
\usepackage{booktabs}

\begin{document}

\title[Article Title]{ScoreCL: Augmentation-Adaptive Contrastive Learning via Score-Matching Function}


\author*[1,2]{\fnm{Jin-Young} \sur{Kim}}\email{seago0828@gmail.com}
\equalcont{These authors contributed equally to this work.}

\author[1]{\fnm{Soonwoo} \sur{Kwon}}\email{swkwon.john@gmail.com}
\equalcont{These authors contributed equally to this work.}

\author[1]{\fnm{Hyojun} \sur{Go}}\email{william@twelvelabs.io}
\equalcont{These authors contributed equally to this work.}

\author[3]{\fnm{Yunsung} \sur{Lee}}\email{sung@wrtn.io}

\author[4]{\fnm{Seungtake} \sur{Choi}}\email{hist0613@naver.com}

\author*[5]{\fnm{Hyun-Gyoon} \sur{Kim}}\email{hyungyoonkim@ajou.ac.kr}

\affil*[1]{\orgname{Twelvelabs}, \orgaddress{\street{Itaewon-ro 27}, \city{Yongsan-gu}, \postcode{04350}, \state{Seoul}, \country{South Korea}}}

\affil*[2]{\orgdiv{Department of Computer Science}, \orgname{Yonsei University}, \orgaddress{\street{Yonsei-ro 50}, \city{Seodaemoon-gu}, \postcode{03722}, \state{Seoul}, \country{South Korea}}}

\affil[3]{\orgname{Wrtn}, \orgaddress{\street{Teheran-ro 2}, \city{Gangnam-gu}, \postcode{06241}, \state{Seoul}, \country{South Korea}}}

\affil[4]{\orgname{Yanolza}, \orgaddress{\street{Teheran-ro 108gil 42}, \city{Gangnam-gu}, \postcode{06176}, \state{Seoul}, \country{South Korea}}}

\affil*[5]{\orgdiv{Department of Financial Engineering}, \orgname{Ajou University}, \orgaddress{\street{World cup-ro 206}, \city{Yeongtong-gu}, \postcode{16499}, \state{Suwon}, \country{South Korea}}}

\abstract{Self-supervised contrastive learning (CL) has achieved state-of-the-art performance in representation learning by minimizing the distance between positive pairs while maximizing that of negative ones. 
Recently, it has been verified that the model learns better representation with diversely augmented positive pairs because they enable the model to be more view-invariant. 
However, only a few studies on CL have considered the difference between augmented views, and have not gone beyond the hand-crafted findings.
In this paper, we first observe that the score-matching function can measure how much data has changed from the original through augmentation. 
With the observed property, every pair in CL can be weighted adaptively by the difference of score values, resulting in boosting the performance.
We show the generality of our method, referred to as ScoreCL, by consistently improving various CL methods, SimCLR, SimSiam, W-MSE, and VICReg, up to 3\%p in image classifcation on CIFAR and ImageNet datasets. 
Moreover, we have conducted exhaustive experiments and ablations, including results on diverse downstream tasks, comparison with possible baselines, and further applications when used with other augmentation methods. 
We hope our exploration will inspire more research in exploiting the score matching for CL.
}




\maketitle
\section{Introduction}
\label{sec:intro}

Self-supervised learning (SSL) of exploiting a large amount of unlabeled data for better representation learning has shown superior results in various computer vision fields, such as object detection~\cite{girshick2014rich, ren2015faster}, semantic segmentation~\cite{long2015fully,he2017mask}, and image classification~\cite{krizhevsky2017imagenet,russakovsky2015imagenet}.
Especially, contrastive learning and its related methods
(CL\footnote{In this paper, we refer to CL as contrastive learning and related methods which model image similarity and dissimilarity (or only similarity) between two or more augmented image views, encompassing siamese networks or joint-embedding methods.})
have shown promising results, not only outperforming previous state-of-the-art self-supervised learning, but also performing comparably to supervised learning~\cite{chen2020simple,chen2021exploring,zbontar2021barlow,bardes2021VICReg,ermolov2021whitening,he2020momentum,chen2020improved,grill2020bootstrap,caron2020unsupervised,li2022twin}. 

The key concept of CL is to encourage the similarity of representations between positive view pairs, which are generated by \textit{data augmentation} from the same image. 
This core scheme remains consistent across several approaches, regardless of incorporation with negative pairs generated from other images~\cite{chen2021exploring,bardes2021VICReg,zbontar2021barlow}.

Therefore, data augmentation has been the major interest of the CL, exploring it for generating better view pairs.
Early works demonstrated the effectiveness of stronger augmentations compared to those typically employed in supervised learning for CL~\cite{chen2020simple,chen2020improved,grill2020bootstrap,wang2022contrastive,xie2022delving}.
However, relying solely on complex augmentations not only fails to ensure the appropriateness of the generated view pairs, in which task-irrelevant information is variously mixed~\cite{tian2020makes, tian2020contrastive}, but  also does not guarantee the pair-wise diversity~\cite{wang2022importance}.
Recently, there have been few attempts to generate diverse view pairs while maintaining task-relevant features and showing its effectiveness: \cite{wang2022importance} showed the effectiveness of asymmetric augmentation keeping a relatively lower variance in one view than another. \cite{peng2022crafting} proposed object-aware center-suppressed sampling. It allows the positive pairs to have common semantics while each has different noises by suppressing sampling from the center of the image.

While advanced augmentation methods have been proposed to vary view pairs, none of them explicitly take into account the \textbf{differences between the pairs}, limited to suboptimal studies that consider all differences between pairs equally. 
To address it, we first formulate the adaptive contrastive loss that focuses on informative pairs, which have substantial differences between the pairs, via weighting. 
The adaptive loss can be applied in conjunction with any augmentation strategies.
However, the challenge lies in the difficulty of measuring the semantic difference between the pair.

We firstly have discovered that the score matching function can estimate the difference. 
The score represents the gradient of the log density with respect to the data, indicating that it is a vector field pointing in the direction where the density increases the most~\cite{hyvarinen2009estimation,hyvarinen2008optimal}. 
However, due to the unknown true distribution, learning the score function remains an intractable challenge~\cite{hyvarinen2009estimation,hyvarinen2008optimal,song2020sliced,vincent2011connection}. 
Denosining score matching (DSM) offers a simple approach to estimating the score by perturbing the data with a noise distribution~\cite{vincent2011connection}. 
In our observations, the strength of augmentation can be accurately measured by evaluating the norm of score values, estimated by DSM, for the augmented images. This remains true even when a combination of multiple augmentations is employed in their generation. 
Furthermore, our findings reveal that the difference between augmented views can be effectively captured by assessing the norm of the score value differences.

Leveraging the observed properties of DSM, we propose a simple but novel CL framework called ``Score-Guided Contrastive Learning'', namely \textbf{ScoreCL}. 
The loss function is designed to attenuate more when the differences between the pairs increase according to their score values. 
To show that our method can be easily applied to existing CL methods regardless of whether they use negative pairs, we empirically validate our method on the benchmark datasets such as CIFAR-10, CIFAR-100~\cite{krizhevsky2009learning}, and ImageNet~\cite{deng2009imagenet,oord2018representation}, achieving up to 3\%p improvements over SimCLR~\cite{chen2020simple}, SimSiam~\cite{chen2021exploring}, W-MSE~\cite{ermolov2021whitening}, and VICReg~\cite{bardes2021VICReg}.
We summarize contributions as follows:

\begin{itemize}
    \item Drawing on empirical evidence of a correlation between score values and the strength of augmentation, we present a novel CL framework, which adaptively focuses on pairs with substantial differences between the pairs. It is not only easily applied to any CL methods, but also to augmentation strategies.
    \item Through extensive experiments, we show that models trained with our method consistently outperform others - even with recent CL methods and augmentation strategies, and a large-scale dataset.
    \item To the best of our knowledge, it is the first work to analyze the property of the score matching function that recognizes the scale of the augmentation. 
\end{itemize}

\DeclareRobustCommand\onedot{\futurelet\@let@token\@onedot}
\def\etal{\emph{et al}\onedot}

\section{Related Work}
\label{sec:relatedwork}

\subsection{Contrastive Learning.}
\label{subsec:CL}

Contrastive Learning (CL) aims to learn transferable representation without labels by using both positive view pairs and negative samples derived from images via stochastic augmentations~\cite{chen2020simple, he2020momentum}. 
Prior work primarily concentrates on designing human-intuitive augmentation strategies, such as random augmentation, center-suppressed sampling, and asymmetric variance augmentation~\cite{tian2020makes,peng2022crafting,wang2022importance}.
However, these strategies neglect the impact of view differences. 
Although some studies briefly explore the effect of varying augmentation scales in CL, they lack a contrastive objective that considers such differences.
Some studies aim to enhance CL's performance by adjusting the weight of components in its loss function.
\cite{song2020multicpc} propose re-weighting contrastive predictive coding loss based on the number of positive samples to tighten the mutual information lower bound.
On the other hand, for view-invariant representations, diverse image augmentation is crucial, striking a balance between task-relevant and unnecessary information to prevent shortcut learning~\cite{tian2020makes}. 
\cite{huang2021towards} introduce a $(\sigma,\delta)$-measure to mathematically quantify the view difference, but don't incorporate it into training, relying on manual changes in augmentation types and intensities for performance comparison. 
\cite{wang2022importance} studies the effect of several asymmetric augmentations used for diverse view generation. 
However, this study also does not go beyond the hand-crafted findings and is vulnerable to the randomness of augmentation. 
Unlike existing work, our distinction is a pair-wise contrastive objective to automatically address view differences, distinguishing it from existing approaches.

\subsection{Score Matching.}
\label{subsec:score}
Score matching is initially presented to train non-normalized statistical models using independent and identically distributed (i.i.d.) samples originating from an unfamiliar data distribution~\cite{hyvarinen2009estimation}. 
The variant of score matching such as sliced score matching or denoising score matching is studied for reducing the extra computation~\cite{vincent2011connection,song2020sliced}.
Due to the property of score matching for regressing the log density of data distribution, it is widely studied from the view of its property~\cite{zhang2021diffusion,gong2021interpreting} or generative model~\cite{song2019generative,song2020score}.
However, to our best knowledge, it is the first work to exploit score matching in contrastive learning by measuring the strength of augmentation.
\definecolor{commentcolor}{RGB}{110,154,155} 

\section{Methodology}
\label{sec:method}
In this section, we first briefly introduce contrastive learning. Then, we formulate the pair-wise adaptive framework. Finally, we present observations and exploit them for the CL.

\subsection{Preliminaries}
\label{subsec:prelim}
\subsubsection{Contrastive representation learning} CL is a general framework for learning encoders so that the distance between positive pairs is close and the distance between negative pairs is far. Positive pairs are typically two views with one image undergoing transformations sampled randomly from the data augmentation pool~\cite{chen2020simple,he2020momentum,grill2020bootstrap}, or they are also defined as data within the same class or sequence~\cite{khosla2020supervised,henaff2020data,oord2018representation}.
Formally, consider a dataset $\mathcal{D}=\{x_i|x_i\in\mathbb{R}^n,i\in\mathcal{I}\}$ where $n$ represent the dimension of data and $\mathcal{I}$ is an index set for data. 
We omit the subscription $i$ for readability.
Let $v$ and $v'$ be the positive pair augmented from $x$ for which we desire to have similar representations, and $z$ be an embedding vector of $v$, i.e., $z=f_\phi(v)$ where $f_\phi$ is an encoder. The similarity between them is obtained by the inner product and it is input to InfoNCE loss for CL:
\begin{equation}
\label{eq:infonce}
  L_{CL}  = - \sum_{i}\log \dfrac{\exp(z^T\cdot z'/\tau)}{\sum_{\gamma\in \Gamma(i)} exp(z^T\cdot z^{\gamma} / \tau)},
\end{equation}
where $\Gamma(i) \equiv \mathcal{I}\setminus \{i\}$, $\tau \in R^+$ is the scalar temperature hyperparameter, and $z^\gamma$ denotes  an embedding for negative pair of the image $z_i$. The denominator encourages the model to distinguish between $x_i$ and samples that are not positive pairs. 
Since this process is expensive in computing resources, a method for learning representations using only numerators (i.e. only positive pairs) has been proposed~\cite{chen2021exploring,zbontar2021barlow,bardes2021VICReg}. 

\subsubsection{Score matching function}
The score matching is introduced to learn a probability density model $q_\theta(x)$
\begin{equation}
    q_\theta(x)=\frac{1}{Z(\theta)}\exp(-E(x;\theta)),
\end{equation}
where $Z(\theta)$ is an intractable partition function, $E$ is an energy function and $\theta$ is parameters of probability density model~\cite{hyvarinen2009estimation}. A score $s:\mathbb{R}^n\rightarrow\mathbb{R}^n$ is called the gradient of the log density with respect to the data, indicating the steepness of the log density. The core principle of score matching is to match the estimated score values $s_\theta(x)=\frac{\partial\log q_\theta(x)}{\partial x}$ to the corresponding score of the true distribution $\frac{\partial\log p(x)}{\partial x}$. The objective function is expected squared error as follows:
\begin{equation}
    \mathbb{E}_{p(x)}\Big[\frac{1}{2}||s_\theta(x)-\frac{\partial\log p(x)}{\partial x}||^2\Big].
\end{equation}
However, due to the unknown true distribution $p$, some methods to regress the above function are proposed~~\cite{hyvarinen2009estimation,hyvarinen2008optimal,song2020sliced,vincent2011connection}. Denoising score matching (DSM) is a simple way to regress the score by perturbing the data with a noise distribution as follows:
\begin{equation}
\label{eq:score}
\mathcal{L}_\sigma=\mathbb{E}_{p_\sigma(x,\tilde{x})}\Big[\frac{1}{2}||s_\theta(\tilde{x})-\frac{\partial\log p_\sigma(\tilde{x}|x)}{\partial \tilde{x}}||^2\Big],
\end{equation}
where $p_\sigma(\tilde{x},x)$ is a joint density $p_\sigma(\tilde{x}|x)p(x)$, $\tilde{x}= x + \sigma \epsilon$ is a perturbed data, noise $\epsilon\sim\mathcal{N}(0,1)$, and $\frac{\partial\log p_\sigma(\tilde{x}|x)}{\partial \tilde{x}}=\frac{1}{\sigma^2}(x-\tilde{x})$.
As shown in~\cite{vincent2011connection}, the objective of DSM is equivalent to that of score matching when the noise is small enough such that $p_\sigma(x)\approx p_{data}(x)$. Several studies have applied these characteristics of DSM to image generation or out-of-distribution detection tasks~\cite{song2019generative, mahmood2020multiscale}.
\subsection{Pair-wise Adaptive Contrastive Learning}
\label{subsec:method}
Recent findings that a positive pair with a wide range of diversity enables representation learning more transferable, more stable converges, and leads to performance enhancement~\cite{tian2020makes,wang2022importance,wang2022contrastive}.
They just focused on how to make the data more diverse, not on how to exploit them more flexibly with respect to the difference in the degree of variant; studying robust learning of $z$ by varying $v$.
However, in this paper, we first formulate the CL objective incorporating how much the $v$ is transformed regardless of the augmentation strategy.
To use the degree of similarity between pairs in a CL objective, with attenuate weight $d(A(v), A(v'))$ where $A(\cdot)$ is a mapping function from images to augmentation scale and $d(\cdot,\cdot)$ is a distance measure, we newly present the adaptive version of InfoNCE loss is represented as follows:
\begin{equation}
\label{eq:score_cl}
\begin{aligned} 
  & L_{A-CL} = \\ &- \sum_{i}\log \dfrac{d(A(v),A(v'))\exp(z^T\cdot z')/\tau)}{\sum_{\gamma\in \Gamma(i)} d(A(v),A(v^{\gamma}))\exp(z^T\cdot z^{\gamma} / \tau)}.
\end{aligned}
\end{equation}
This proposed adaptive loss puts more weight on learning pairs with a large difference in the augmentation strength. 

Some work tried a similar approach to address the hard negative (difficult to distinguish negative pairs) or hard positive (hard to find similarity) pairs~\cite{lee2022r,robinson2020contrastive}, but they simply use the fixed constant as the weight $A(\cdot)$ for similarity of pairs, yielding inflexible CL objective. 
On the other hand, for estimating the function $A(\cdot)$, existing works~\cite{huang2021towards,wang2022importance} only focus on finding the impact of differences in augmentation scale in a hand-crafted manner. 
Furthermore, these hand-crafted augmentation scales are globally fixed, resulting in the lack of pair-wise estimation and sub-optimal. 
To overcome this limitation, we study a method to measure the scales automatically with an observation on score matching in the following section.

\subsection{Observation on Score Matching}
\begin{figure}[t]
    \centering
    \includegraphics[width=0.8\columnwidth]{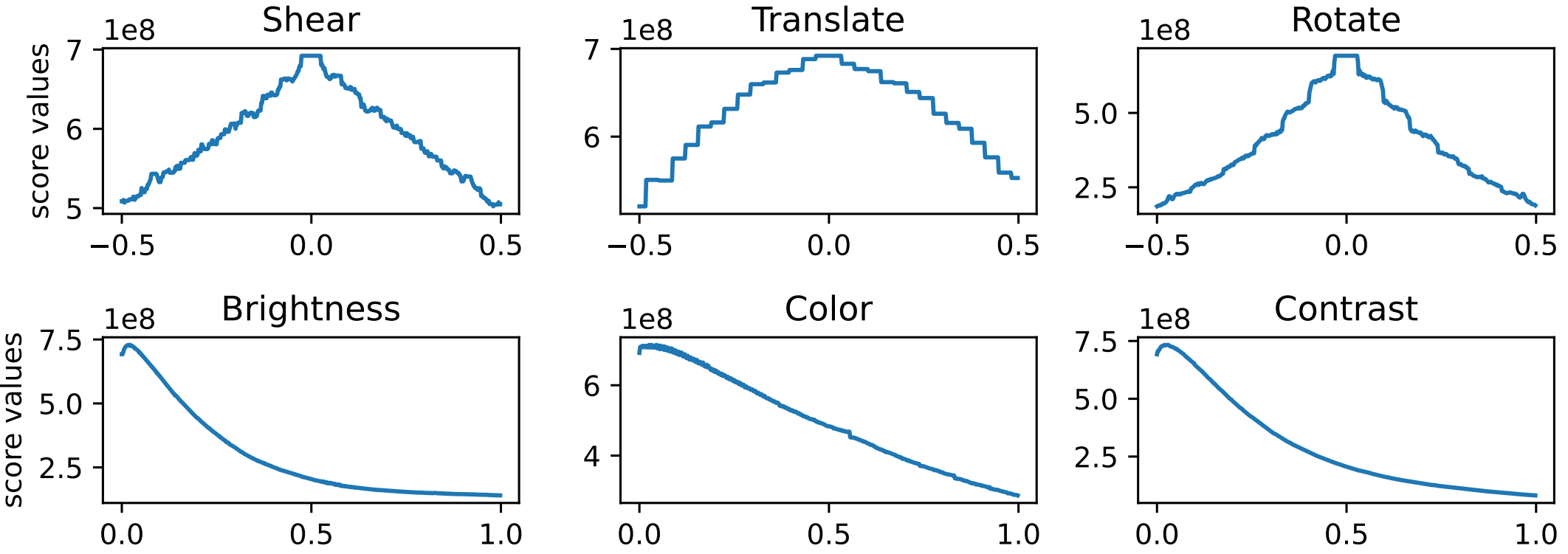}
    \caption{The score values - magnitude of augmentation graph for each transform which are sampled from RandAugment~\cite{cubuk2020randaugment}.
    }
    \label{subfig:aug_score}
\end{figure}

\begin{figure}
\centering
    \includegraphics[width=\columnwidth]{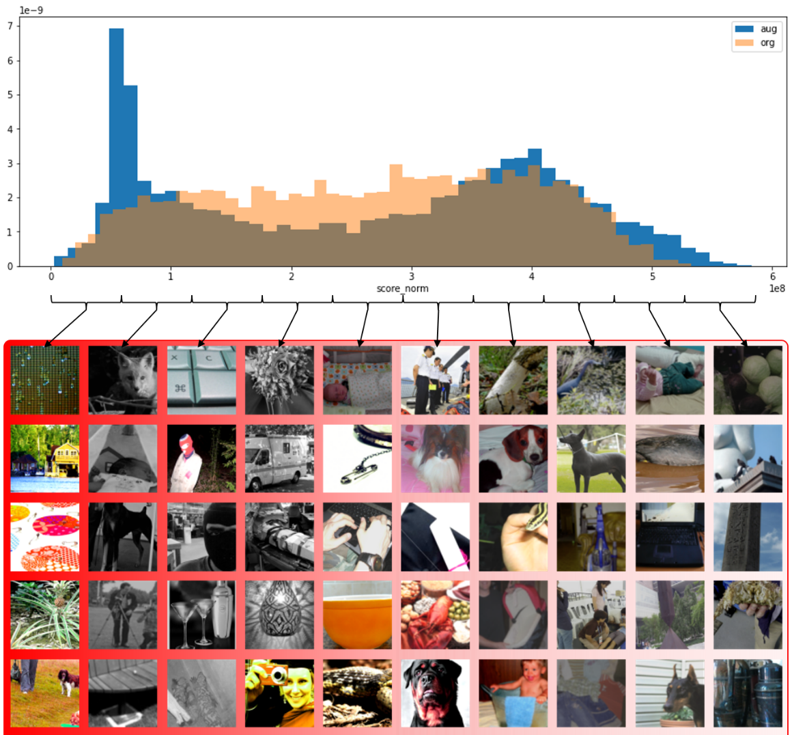}
    \caption{The histogram of score values and sample images in the binning range. We confirm that, unlike the distribution of the score values of the original image, that of augmented images has a peak. Through the qualitative analysis, we confirm that the transformed images with high intensity of augmentation (especially, color-related transform) have low score values as shown in the left two columns. 
    }
    \label{fig:score_example}
\end{figure}

\begin{figure}
    \centering
    \includegraphics[width=\columnwidth]{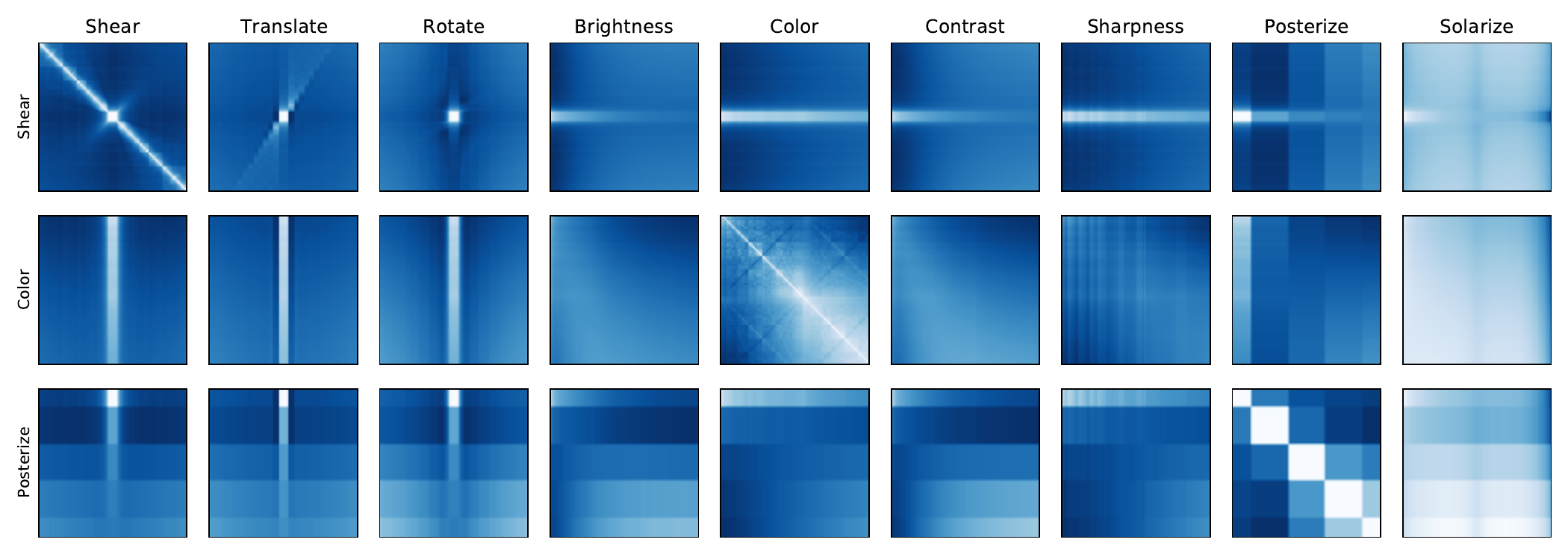}
    \caption{The observation that the difference of score value is related to that of augmentation-scale. Note that the ``Shear", and ``Translate" transforms have negative directions, so we align the original images (i.e. zero magnitudes) in the middle across the axis. We can find that the difference in score values is smaller as the degree of transforms is closer. For example, when both views are transformed with ``Color" (at the second row and fifth column), if the magnitude increases to the same degree, the difference between score values is low, and if the difference in magnitude is large, the difference in score is also increased.
    }
    \label{fig:2aug_2img}
\end{figure}
\begin{figure}[t]
    \centering
    \includegraphics[width=0.8\columnwidth]{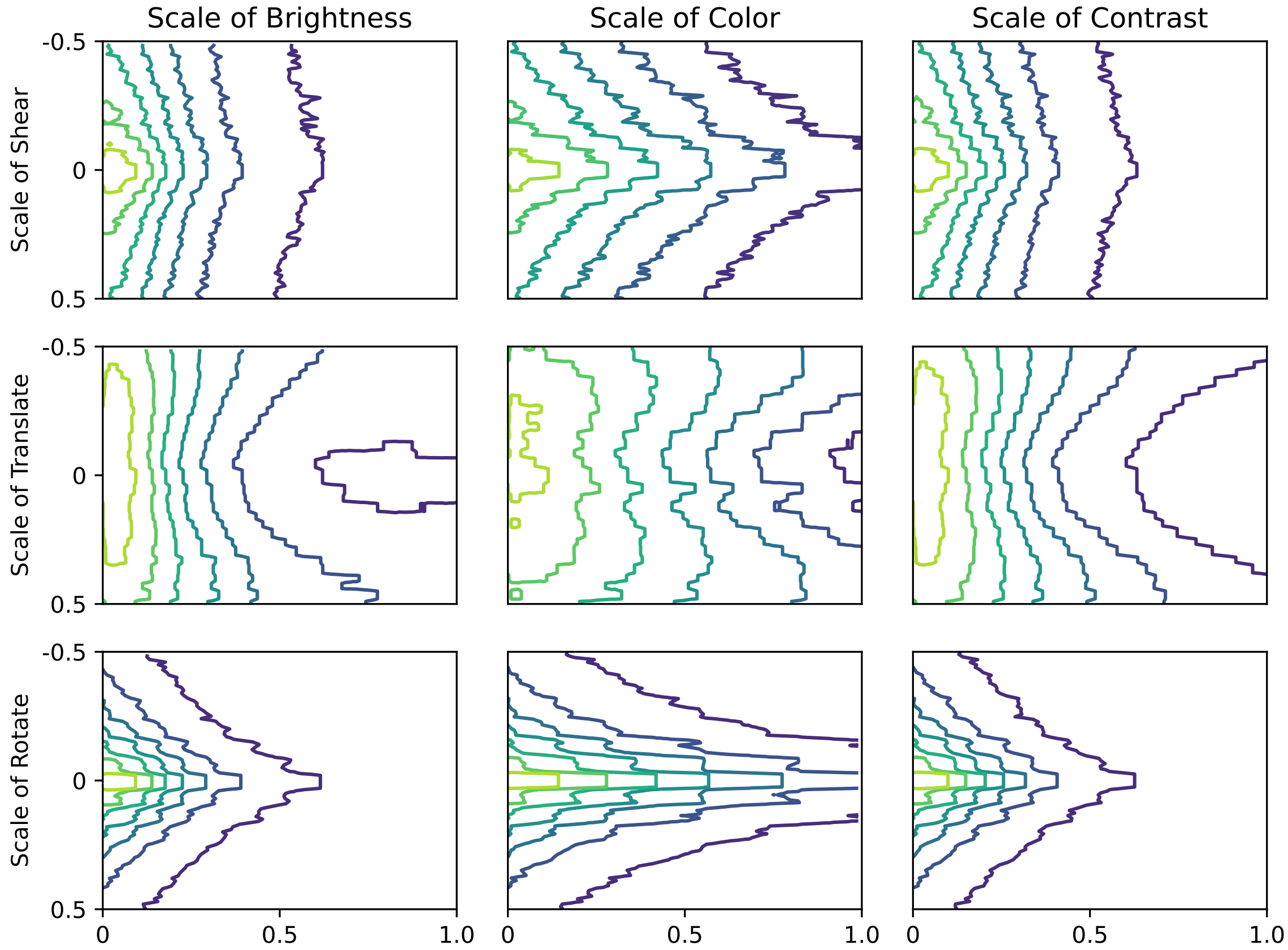}
    \caption{Contour map of score values when two augmentations are applied to one image. Each axis corresponds to the augmentation scale. We can confirm the non-linear relation between score values and augmentation scale when more than two transforms are applied to one images.
    }
    \label{subfig:2aug_1img}
\end{figure}

Our work aims to make CL loss adaptively use the difference in the augmentation scale of views.
Note that the score value can be viewed as the output of a function that measures the level of noise embedded in the input image. 
Here we rely on two results: 1) the gradient of the logarithm of the noisy signal density can be expressed as solutions to remove additive Gaussian nosie~\cite{kadkhodaie2021stochastic} and 2) the Gaussian noise-based image degradation plays a similar role in diffusion models with other augmentations, even completely deterministic degradation, e.g., blur masking and more~\cite{bansal2022cold,sohl2015deep}. 
Referring to them, we hypothesize that the score values for the augmented samples would be related to the corresponding noise level so as to be a degree of transformation.

To ensure the possibility of using score values for pair-wise adaptive contrastive learning, the three observations have to be verified. 
Let $v^{\{a\}}$ be the augmented view $v$ with transform set $\{a\}$, $\mathrm{P}(a)$ be the scale of the augmentation $a$, and $x\sim y$ denotes that $x$ and $y$ are correlated.
First, the score value should correlate with the strength of a single augmentation, only then can we design CL considering the difference in view pairs: $s_\theta(v^a)\sim \mathrm{P}(a)$. 
Note that the score values are in $\mathbb{R}^n$, so we analyze the magnitude of them.
Second, the score value must correlate with the strength of at least two augmentations to infer the correlation for multiple augmentations inductively: $s_\theta(v^{\{a,b,c,\dots\}})\sim\psi(\mathrm{P}(a),\mathrm{P}(b),\mathrm{P}(c),\dots)$ where $\psi$ is aggregate function to measure the augmentation scale when multiple transforms are applied.
Third, the gap in score values should be correlated with the gap in augmentation strength so as to assign a pair-wise weight based on the score value of each view in CL: $\Delta(s_\theta(v^{\{a\}}), s_\theta(v^{\{b\}}))\sim\Delta(\mathrm{P}(a),\mathrm{P}(b))$.
We will discuss further details of the analysis below.

\subsubsection{Analysis on score values.} As illustrated in Figs. \ref{subfig:aug_score}-\ref{subfig:2aug_1img}, we analyze the relationship between score values and augmentation strength. Since we use DSM~\cite{vincent2011connection} whose output is the gradient of the log density of data by regressing it with perturbing data 
 as shonw in equation \ref{eq:score}, i.e., $\frac{(x-\tilde{x})}{\sigma^2}=-\frac{\epsilon}{\sigma}$ where $\tilde{x}=x+\sigma\epsilon$ and $\epsilon\sim\mathcal{N}(0, 1)$ such that the large degree of deformation ($\sigma$) of the image results in the small absolute score value.

\textbf{1) Score values have a correlation with the strength of single augmentation.}
Though the views are augmented with various transformation, we first analyze how the score values of augmented samples change according to the strength of the augmentation applied to the image.
Figure~\ref{subfig:aug_score} shows that the score value tends to decrease as the strength of augmentation increases. 
For the qualitative analysis, we plot the Fig. \ref{fig:score_example} which shows the histogram of score values obtained from ImageNet dataset~\cite{deng2009imagenet} with RandAugment~\cite{cubuk2020randaugment}.
The images in the left two columns with small score values are deformed with color-related augmentation such as color jittering or grayscale.
From these, we confirm that score values and the augmentation strength are correlated.

\textbf{2) Gap of score values have a correlation with the strength gap between two different augmentations.}
For using the score values in the CL, they should also be related to augmentations applied to make two views.
Therefore, we investigate the difference in score values of the two views that are produced by various augmentation strengths from the image.
At first, each view is transformed with distinct augmentation in various scale.
Then the score values of them is obtained and subtracted.
As shown in Fig.~\ref{fig:2aug_2img}, the difference of score values is small if the scale of augmentation is similar, and vice versa.
For example, the image in the first row and the fifth column is a heatmap of the difference in score values from two images which are augmented with `Shear' and `Color' augmentation, respectively.
As we mentioned, note that the magnitude of `Shear' augmentation is aligned in the middle across the y-axis. 
In the middle of the y-axis, where `Shear' is hardly applied, the difference in score values increases according to the strength of `Color', whereas the opposite trend appears after `Shear' is applied to some extent. 
It can be interpreted that the distance from the true distribution is rather far, so the difference between the two gradients (score values) is recognized as small.
Otherwise, the case where the relationship between the difference in score value and the difference in augmentation strength can be most easily confirmed is in the third row and eighth column (i.e., when the ``Posterize" transform is applied to each of the two images).
Through this, we conjecture that the difference of score values of the two views is correlated with the difference of the scale of augmentation.

\textbf{3) Score values have a non-linear correlation with the strength of a combination of multiple augmentations.}
From the above observations, one might design the adaptive CL objective by naively utilizing the difference in the scale of applied augmentation instead of using the score values.
However, multiple augmentation methods can be sequentially applied to generate views and it is difficult to estimate the strength of these composited augmentations by the simple linear combination of each augmentation strength.
Therefore, we analyze the expressivity of the score matching function when two augmentations are applied and illustrate this in Fig.~\ref{subfig:2aug_1img}.
The results show that the score value is expressed as a non-linear combination of their intensities.
Besides, unlike the implementation of two augmentations, it is further difficult to analyze from more complex combinations of several augmentations, such that it shows the need for much simpler methods like ours.

\begin{figure}
    \centering
    \includegraphics[width=0.8\columnwidth]{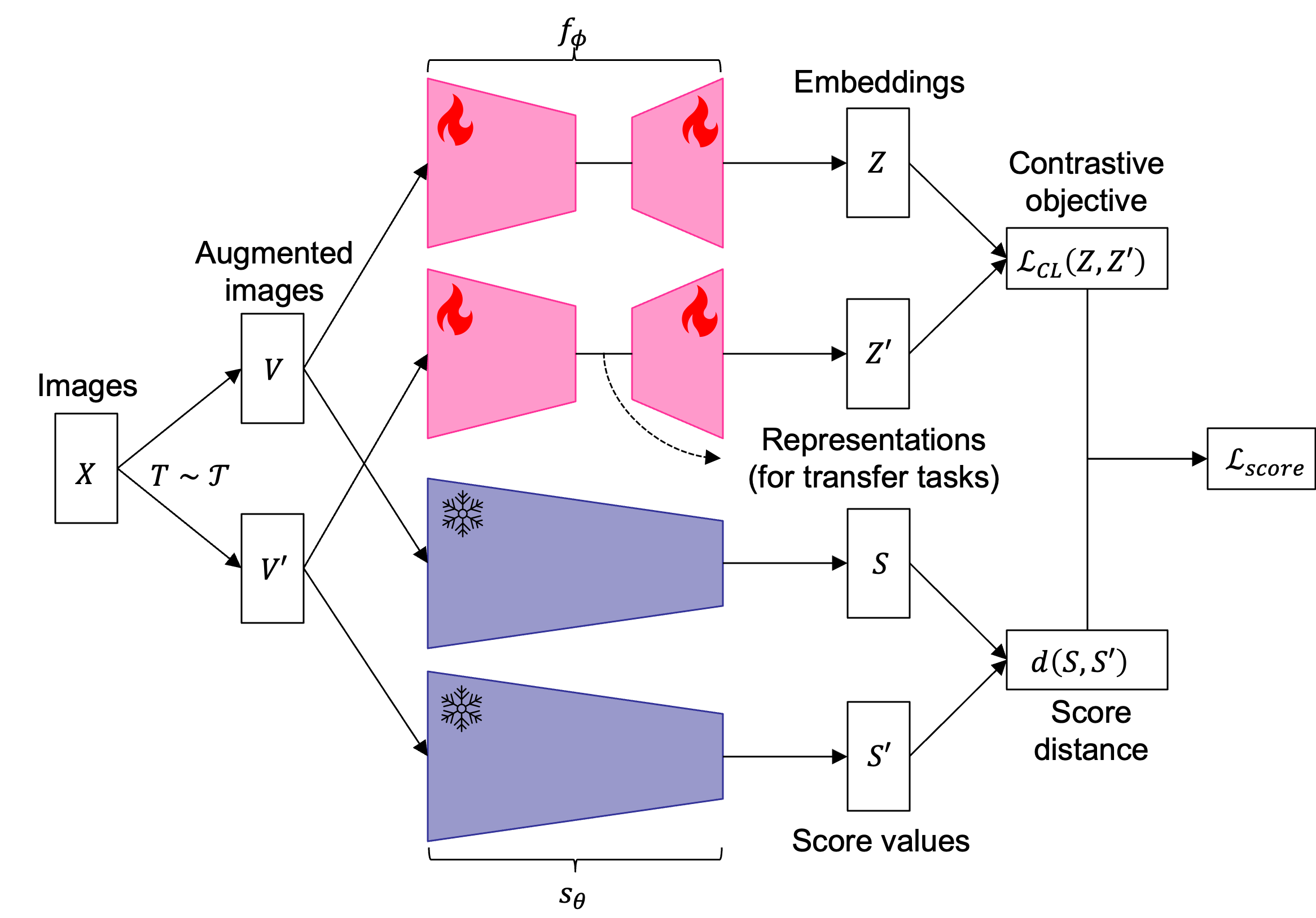}
    \caption{The architecture of ScoreCL with score matching function $s_\theta$. The red diagram acts as the original CL model and the blue figure represents score guiding; it can represent the existing CL method with $d(S,S')=1$. Note that the $s_\theta$ is trained before CL so as to prevent the gradient flow through score matching. 
    }
    \label{fig:architecture}
\end{figure}

\subsection{Score-Guided Contrastive Learning}

From the above analysis, the score matching function can be used to estimate the difference in the strength of transforms applied to each view, we propose a score-guided CL that learns adaptively to attenuate hard positives by utilizing the characteristic of the score values.
We set $A(\cdot)$ in equation \ref{eq:score_cl} as $s_\theta(\cdot)$, which is illustrated in Fig. \ref{fig:architecture}. 
The PyTorch-style pseudocode is shown in Algorithm~\ref{alg:scoreCL}.
\begin{algorithm}[t]
\caption{PyTorch-style pseudocode for ScoreCL}
\label{alg:scoreCL} 
\begin{algorithmic}[1] 
\STATE {\bfseries Input:} An encoder network $f_\phi$, a score matching $s_\theta$, contrative objective $\mathcal{L}_{CL}$, and distance measure $d$
\STATE for $x$ in loader:
    \STATE \hskip1.0em $v_1$, $v_2$ = augment$(x)$
    \STATE \hskip1.0em\textcolor{commentcolor}{\# compute embeddings}
    \STATE \hskip1.0em$z_1, z_2=f_\phi(v_1), f_\phi(v_2)$
    \STATE \hskip1.0em\textcolor{commentcolor}{\# compute score values}
    \STATE \hskip1.0em$s_1, s_2=s_\theta(v_1), s_\theta(v_2)$
    \STATE \hskip1.0em\textcolor{commentcolor}{\# obtain loss for contrastive learning}
    \STATE \hskip1.0emcl\_loss = $\mathcal{L}_{CL}(z_1,z_2)$
    \STATE \hskip1.0em\textcolor{commentcolor}{\# obtain distance between scores from views}
    \STATE \hskip1.0emscore\_dist = $d(s_1,s_2)$
    \STATE \hskip1.0em\textcolor{commentcolor}{\# scoreCL}
    \STATE \hskip1.0emloss = (score\_dist.detach() * cl\_loss).mean(0)
    \STATE \hskip1.0em\textcolor{commentcolor}{\# optimizatin step}
    \STATE \hskip1.0emloss.backward()
    \STATE \hskip1.0emoptimizer.step()

\end{algorithmic}
\end{algorithm}
The distance $d(A(v),A(v'))$ is set as the L1 norm, and we confirm that even when $d$ is set to the L2 norm, results similar to those analyzed above are obtained.
To verify the generality of our approach to existing methods, we select the four different types of method as presented in~\cite{bardes2021VICReg}: SimCLR (Contrastive learning), SimSiam (Distillation methods), W-MSE (Information maximization methods), and VICReg (Joint embedding).
Note that our method can be applied to every method orthogonally to the usage of negative samples and improve the baselines consistently. 
The modified loss functions for each method are as follows:
\begin{itemize}
    \item \textbf{SimCLR}~\cite{chen2020simple}:
    \begin{equation}
        -\log\frac{d(s_\theta(v),s_\theta(v'))\exp(\text{sim}(z,z')/\tau)}{\sum_{\gamma\in \Gamma(i)} d(s_\theta(v),s_\theta(v^\gamma))\exp(z\cdot z^{\gamma} / \tau)}.
    \end{equation}
    
    \item \textbf{SimSiam}~\cite{chen2021exploring}:
    \begin{equation}
        \frac{1}{2}d(s_\theta(v),s_\theta(v))(\mathcal{D}(p,z')+\mathcal{D}(p',z)),
    \end{equation}
    where $p$ is a representation from $z$ passing through prediction MLP head and $\mathcal{D}(x,y)=-\frac{x}{||x||_2}\cdot\frac{y}{||y||_2}$ which is a cosine similarity to measure the similarity.
    
    \item \textbf{W-MSE}~\cite{ermolov2021whitening}:
    \begin{equation}
        \frac{2}{Nm(m-1)}\Sigma d(s_\theta(v),s_\theta(v'))dist(w,w'),
    \end{equation}
    where $N$ is the number of given original images, $k$ is a batch size, $m=K/N$, and $w$ is a whitened vector from $z$. $dist$ is a distance measure with MSE between normalized vectors.
    
    \item \textbf{VICReg}~\cite{bardes2021VICReg}:\\
    \begin{equation}
        \lambda A_{score}(Z,Z')+\mu[B(Z)+B(Z')]+\eta[C(Z)+C(Z')],
    \end{equation}
    where $\lambda$, $\mu$, and $\eta$ are hyper-parameters balancing the importance of each term. We denote that $Z$ and $Z'$ are the set of $z$ and $z'$, respectively. $A_{score}(Z,Z')=\frac{1}{n}\Sigma_id(s_\theta(v),s_\theta(v'))||z-z'||_2^2$ is invariance criterion. $B(Z)=\frac{1}{d}\Sigma_{j=1}^d\max(0,1-S(z^j,\epsilon))$ is variance regularization where $S(x,\epsilon)=\sqrt{\text{Var}(x)+\epsilon}$, and $z^j$ is the vector composed of each value at dimension $j$ in $Z$. $C(Z)=\frac{1}{d}\Sigma_{i\neq j}|c(Z)|^2_{i,j}$ is covariance regularization, where $c(Z)$ is a covariance matrix of $Z$.
\end{itemize}
\subsubsection{Training score matching model.} Before learning representation in CL, we first pre-train the score matching using equation \ref{eq:score}. Following~\cite{song2019generative}, since $\mathcal{L}_\sigma$ relies on the scale of $\sigma$, we use unified objective with all $\{\sigma_k\}_{k=1}^L$, not just one $\sigma$ as follows:
\begin{equation}
  \mathcal{L}_s=\frac{1}{L}\Sigma_{k=1}^L\xi(\sigma_k)\mathcal{L}_{\sigma_k},
\end{equation}
where $\xi(\sigma)=\sigma^2$ to derive $||\sigma s_\theta(\tilde{x})||_2 \propto 1$. In fact, our method increases training costs, but it is minor since the score matching function and CL are trained separately. We discuss the details of training costs through the empirical analysis.

\section{Experiments}
\label{sec:experiments}
In this section, we verify the ScoreCL from various perspectives. 
At first, we evaluate the ScoreCL on linear and k-NN classifications on top of the frozen representations trained on various datasets such as ImageNet, CIFAR-100, and CIFAR-10.
Since the general evaluation protocol for unsupervised representation learning is the performance on the downstream task~\cite{chen2021exploring, bardes2021VICReg}, the transfer learning for image classification and fine-tuning on object detection is conducted.
To better highlight our technical contributions, we conduct additional experiments: other weighting schemes, augmentation strategies, false positive pair, and batch size.
The code will be released after acceptance.

\label{sec:setup}
\noindent\textbf{Datasets and CL Models.} To verify the consistent superiority of the proposed method, experiments have been conducted on various datasets and existing CL models. 
We use well-known benchmark datasets such as CIFAR-10, CIFAR-100~\cite{krizhevsky2009learning}\footnote{\url{https://www.cs.toronto.edu/~kriz/cifar.html}}, ImageNet-100~\cite{oord2018representation}, and ImageNet-1K~\cite{deng2009imagenet}\footnote{\url{https://www.image-net.org/}} for training CL models such as SimCLR~\cite{chen2020simple}, SimSiam~\cite{chen2021exploring}, W-MSE~\cite{ermolov2021whitening}, and VICReg~\cite{bardes2021VICReg}.
For the augmentation strategy (C,C), We follow the settings of~\cite{chen2020simple}: we extract crops with a random size from 0.2 to 1.0 of the original size and also apply horizontal mirroring with probability 0.5. 
Color jittering with configuration (0.4, 0.4, 0.4, 0.1) with probability 0.8 and grayscaling with probability 0.2 are applied. 
For ImageNet-100, we add Gaussian blurring with a probability of 0.5.
For a SimCLR and SimSiam, we use the SGD optimizer with momentum 0.9~\cite{sutskever2013importance}. 
For W-MSE and VICReg, we follow the official settings: Adam and LARS optimizers~\cite{kingma2014adam,you2019large}, respectively. 
Specifically, on CIFAR-10 and CIFAR-100 datasets, for SimCLR, we train CL for 1,000 epochs with a learning rate 0.5, weight decay 0.0001, and temperature 0.5; for SimSiam, 1,000 epochs with a learning rate 0.06 and weight decay 0.0005; for W-MSE, 1,000 epochs with learning rate 0.001 and weight decay $10^{-6}$. 
On ImageNet datasets, for SimCLR, the model learns for 200 epochs with learning rate 0.5, weight decay 0.0001, and temperature 0.5; for SimSiam, 200 epochs with a learning rate 0.1 and weight decay 0.0001. 
Also, we use a cosine learning rate decay with 10 epochs warm-up for all experiments. The embedding size and hyperparameters configuration is set as that of the original paper. 
All experiments were conducted on a single NVIDIA A100 GPU.


\begin{table}[t]
\centering
\resizebox{0.8\columnwidth}{!}{
\begin{tabular}{c|cc|cc|cc}
\toprule
Classifier  & \multicolumn{2}{c|}{SimCLR} & \multicolumn{2}{c|}{Simsiam} & \multicolumn{2}{c}{VICReg} \\
        & base             & ScoreCL             & base              & ScoreCL & base              & ScoreCL             \\ \midrule
k-NN  & 42.62                 &  \textbf{45.59(+2.97)}                 &  54.66               &   \textbf{54.98(+0.32)} &  55.08               &   \textbf{57.80(+2.72)}                \\
Linear & 56.98                 &  \textbf{59.21(+2.23)}                 &          58.13         &   \textbf{59.98(+1.85)}&          59.58         &   \textbf{61.86(+2.28)}           \\ \bottomrule
\end{tabular}
 }
\caption{The classification accuracy for each classifier on ImageNet-1K dataset with ResNet-50 encoder.}
\label{table:main_results_img}
\end{table}

\begin{table}
\centering
\resizebox{0.8\columnwidth}{!}{
\begin{tabular}{c|cc|cc|cc}
\toprule
Method & \multicolumn{2}{c|}{ImageNet-100} & \multicolumn{2}{c|}{CIFAR-10} & \multicolumn{2}{c}{CIFAR-100} \\
        & base    & ScoreCL         & base    & ScoreCL            & base    & ScoreCL            \\ \midrule
SimCLR & 69.24      & \textbf{72.26(+3.02)} & 90.28      & \textbf{91.01(+0.73)}   & 60.11      & \textbf{62.34(+2.23)}   \\
SimSiam& 73.24      & \textbf{74.18(+0.94)} & 90.27      & \textbf{90.80(+0.53)}   & 63.15      & \textbf{64.55(+1.40)}   \\
W-MSE & -      & -  & 90.06      & \textbf{91.35(+1.29)}   & 56.69      & \textbf{56.94(+0.25)}   \\
VICReg & 70.24      & \textbf{71.44(+1.20)} & 88.94      & \textbf{89.49(+0.55)}   & 59.95      & \textbf{61.53(+1.58)}   \\ \bottomrule
\end{tabular}
}

\caption{The classification accuracy of a 5-NN classifier for different CL and datasets. We use ResNet-50 encoder for ImageNet-100 and ResNet-18 for others. This demonstrates the generalizability of ScoreCL, as it enhances the performance of all baseline CL methods. We could not report the results on W-MSE due to the OOM issue.}
\label{table:main_results_cifar}
\end{table}


\begin{table}[t]
\centering
\resizebox{0.8\columnwidth}{!}{
\begin{tabular}{c|cc|cc}
\toprule
Dataset   & Method  & Architecture & base  & ScoreCL        \\ \midrule
ImageNet-100  & SimCLR  & ResNet-101   & 70.14 & \textbf{72.90(+2.76)} \\
          & SimSiam & ResNet-101   & 74.02 & \textbf{75.04(+1.02)} \\
          & VICReg  & ResNet-101   & 71.56 & \textbf{72.24(+0.68)} \\ \midrule
CIFAR-100 & SimCLR  & ViT-S/4      & 58.49 & \textbf{60.91(+2.42)} \\
          & SimCLR  & ViT-S/8      & 55.51 & \textbf{58.01(+2.50)} \\
          & SimCLR  & ViT-S/16     & 47.11 & \textbf{50.44(+3.33)} \\
          & SimCLR  & ViT-B/8      & 55.87 & \textbf{59.17(+3.30)} \\
          & SimCLR  & ViT-B/16     & 47.11 & \textbf{48.56(+1.45)} \\ \bottomrule
\end{tabular}
}
\caption{The classification accuracy of a 5-NN classifier for different encoders.}
\label{table:main_results_arch}
\end{table}

\begin{table}
\centering
\resizebox{0.8\columnwidth}{!}{
\begin{tabular}{cc|cccccc}
\toprule
      \multicolumn{2}{c|}{Dataset}
      & STL10           &Food& Flowers         & Cars            & Aircraft        & DTD  \\ \midrule
\multicolumn{1}{c|}{k-NN}&base  & 80.56      & \textbf{46.31}   & 61.60          & 16.03          & \textbf{21.57} & 54.57          \\
\multicolumn{1}{c|}{}&ScoreCL & \textbf{83.04} & 45.97&\textbf{64.55} & \textbf{16.23} & 21.48          & \textbf{56.09} \\ \midrule
\multicolumn{1}{c|}{Linear}&base & 87.79        & 59.07 & 64.04          & \textbf{26.96}          & 25.33          & 55.85\\
\multicolumn{1}{c|}{}&ScoreCL & \textbf{88.79} & \textbf{59.34}&\textbf{66.03} & 25.46 & \textbf{26.77} & \textbf{57.28}

\\ \bottomrule
\end{tabular}
}
\caption{Downstream task performance on object classification. We report average per-class accuracy for Aircraft and Flowers and top-1 for others. Note that we tune the hyperparameters of linear classifiers only with learning rates in $\{10,30,50,70,90\}$. Thus, the results of the linear classifier do not match those in the previous works, but ours almost always outperforms the baseline.}
\label{table:downstream_cls}
\end{table}

\begin{table}[t]
\centering
\resizebox{0.8\columnwidth}{!}{
\begin{tabular}{c|ccc|ccc|ccc}
\toprule
      Method
      & \multicolumn{3}{c|}{VOC07+12}                                          & \multicolumn{3}{c|}{COCO detection}                                                             & \multicolumn{3}{c}{COCO segmentation}                                                              \\
      SimSiam & $\text{AP}_\text{all}$ & $\text{AP}_\text{50}$ & $\text{AP}_\text{75}$ & 
      $\text{AP}^\text{bb}$ & $\text{AP}_\text{50}^\text{bb}$ & $\text{AP}_\text{75}^\text{bb}$ & $\text{AP}^\text{mk}$ & $\text{AP}_\text{50}^\text{mk}$ & $\text{AP}_\text{75}^\text{mk}$ \\ 
      
      \midrule
base  & 53.87                & 79.85               & 59.51               & 39.15               & 59.03                         & 42.82                         & 34.33               & 55.60                         & 36.63                         \\
ScoreCL & \textbf{53.99}       & \textbf{80.21}      & \textbf{59.69}      & \textbf{39.92}      & \textbf{59.67}                & \textbf{43.13}                & \textbf{34.84}      & \textbf{56.30}                & \textbf{37.18}                \\ \bottomrule
\end{tabular}
}
\caption{The results of downstream tasks on object detection on VOC07+12 using Faster R-CNN~\cite{ren2015faster} and in the detection and instance segmentation task on COCO using Mask R-CNN~\cite{he2017mask}. 
The experiment is done by using MoCo~\cite{he2020momentum} official implementation based on Detectron2~\cite{detectron2}.}
\label{table:downstream_detection}
\end{table}

\subsection{Quantitative Evaluation}
\noindent\textbf{Image Classification.} 
Table \ref{table:main_results_img} shows the image classification accuracy improvement on a large scale dataset like ImageNet when using our method for various CL methods.
Table \ref{table:main_results_cifar} shows the results on diverse datasets.
Notably, ScoreCL outperforms its competitors, showing the advantages of adaptively considering the view pairs during contrastive learning regardless of the dataset.
To verify our method on the different encoder, we conducted experiments using various architectures, including the Vision Transformer (ViT;~\citealt{dosovitskiy2021an}). 
Especially, as shown in Table \ref{table:main_results_arch}, we observed that even smaller models trained with ScoreCL outperform bigger baselines. Specifically, on SimCLR and SimSiam, ResNet-50 with ScoreCL achieves 72.26 and 74.18, which are higher than the 70.14 and 74.02 of ResNet-101.
\begin{table}[t]
\centering
\resizebox{0.8\columnwidth}{!}{
\begin{tabular}{c|cccccc}
\toprule
Model   & Base  & Random & Pixel & Saliency & LPIPS & ScoreCL \\ \midrule
SimCLR  & 60.11 & 60.08 & 59.39 & 60.59    & 60.11 & \textbf{62.34} \\
Simsiam & 63.15 & 56.37   & 63.02 & 63.76      & 61.99 & \textbf{64.55} \\ \bottomrule
\end{tabular}
}
\caption{The comparison results with other weighted CL baselines on the CIFAR-100 dataset.}
\label{table:ab_wcl}
\end{table}

\begin{table}[t]
\centering
\resizebox{0.7\columnwidth}{!}{
\begin{tabular}{c|cc|cc|cc}
\toprule
Method  & \multicolumn{2}{c|}{(C,C)} & \multicolumn{2}{c|}{(C$^+$,C$^+$)} & \multicolumn{2}{c}{(C,R)} \\
        & base    & ScoreCL          & base    & ScoreCL          & base         & ScoreCL              \\ \midrule
SimCLR  & 60.11   & \textbf{62.34}   & 57.80   & \textbf{60.30}   & 63.06        & \textbf{65.26}       \\
SimSiam & 63.15   & \textbf{64.55}   & 59.63   & \textbf{60.76}   & 66.91        & \textbf{67.70}       \\ \bottomrule
\end{tabular}
}
\caption{The ablation study on different augmentation processes with CIFAR-100 dataset. 
The `C' and `C$^+$' represent the customized augmentation with details in the appendix and the `R' means RandAugment.
$(X,Y)$ shows that $X$ augmentation is applied for one view and $Y$ for the other one.}

\label{table:ab_aug}
\end{table}
\begin{table}[t]
\centering
\resizebox{0.7\columnwidth}{!}{
\begin{tabular}{c|cc|cc}
\toprule
Augmentation       & \multicolumn{2}{c|}{CIFAR-10} & \multicolumn{2}{c}{CIFAR-100} \\
       & base          & ScoreCL         & base      & ScoreCL             \\ \midrule
RandomCrop & 90.40         & \textbf{90.94}         & 63.86     & \textbf{64.03}    \\
ContrastiveCrop & 90.72         & \textbf{90.99}         & 64.47     & \textbf{64.56}    \\ \bottomrule
\end{tabular}
}
\caption{Linear classification results on SimSiam with ResNet-18 for different datasets. We set the experimental setting such as linear evaluation protocol or augmentation strategies as in \cite{peng2022crafting}.}
\label{table:ab_fp}
\end{table}

\noindent\textbf{Downstream Tasks.}
To evaluate the generalizability of the learned representation, we conduct transfer learning on a variety of different fine-grained datasets. 
For the image classification task, we follow the k-NN and linear evaluation protocols on the benchmark datasets such as STL10\footnote{\url{https://cs.stanford.edu/~acoates/stl10/}}, Food101\footnote{\url{https://data.vision.ee.ethz.ch/cvl/datasets_extra/food-101/}}, Flowers102\footnote{\url{https://www.robots.ox.ac.uk/~vgg/data/flowers/102/}}, StanfordCars\footnote{\url{https://ai.stanford.edu/jkrause/cars/car_dataset.html}}, Aircraft\footnote{\url{https://www.robots.ox.ac.uk/vgg/data/fgvc-aircraft/}}, and DTD\footnote{\url{https://www.robots.ox.ac.uk/vgg/data/dtd/}}~\cite{coates2011analysis,bossard2014food,nilsback2008automated,krause20133d,maji13fine-grained,cimpoi14describing}. 
For linear evaluation, we train a classifier on the top of frozen representations of ResNet-50 trained in SimSiam as done in the previous works~\cite{grill2020bootstrap, lee2022r}. 
The results are in Table \ref{table:downstream_cls}.
Note that ScoreCL improves baseline in four out of six datasets in k-NN classification and five out of six datasets in linear evaluation, especially showing superior k-NN classification performance on Flowers102 (+2.9\%p), and DTD (+1.5\%p). 

Also, we evaluate the trained representation by object detection and instance segmentation task. Following the setup in~\cite{he2020momentum}, we use the VOC07+12~\cite{everingham2010pascal} and COCO~\cite{lin2014microsoft} datasets. The experimental results in Table \ref{table:downstream_detection} show that ScoreCL makes the existing CL methods enhanced for localization tasks as well.

\subsection{Extensive Analysis}
\label{sec:ab}

\noindent\textbf{Comparison with Adaptive CL Baselines.} As shown in Fig.~\ref{subfig:2aug_1img}, for multiple augmentation scenarios, it is difficult to leverage the naive augmentation scale since the method to combine each scale is not uncovered, for which we adopt score values for better estimating it. 
To highlight the advantages of using score values, here we introduce some baselines for estimating the augmentation scales. 
Specifically, we replace the weight $A(\cdot)$ with the following metrics: random, pixel-wise distance, saliency map (reflecting the task-relevant feature), and LPIPS (measuring the perceptual similarity)~\cite{zhang2018unreasonable}. 
In the case of LPIPS, for example, we adaptively penalize the CL objective with the distance between the LPIPS values of each view.
The results are presented in Table \ref{table:ab_wcl}, showing that ScoreCL improves performance consistently.

\noindent\textbf{Comparison over Various Augmentations.} To show that our method could boost performances regardless of augmentation strategies, we enforce two views to be different by applying different augmentation methods to each view as in~\cite{wang2022importance}.
Furthermore, to compare with the naive strategy that increased the augmentation scale, we conduct experiments with much stronger augmentation (C$^+$, C$^+$) by applying further color jittering, gray-scaling, and cropping.
In the (C$^+$,C$^+$) strategy, we apply further image cropping, color jittering and grayscaling: crop with a random size from 0.1 to 1.0, color jittering with configuration (0.8, 0.8, 0.8, 0.2) with probability 0.8, and grayscaling with probability 0.4.
Table \ref{table:ab_aug} shows the results of proving that the ScoreCL adaptively penalizes the contrastive objective for any augmentation strategies. 
Besides, the result manifest that our method further improves performance with stronger augmentation, while the baseline even degrades performance.
Therefore, applying stronger augmentation to both views does not always increase the performance and does not have a similar effect to our method.

\begin{table}[]
\centering
\resizebox{0.7\columnwidth}{!}{
\begin{tabular}{c|c|c|cc}
\toprule
Method  & Arch.           & Weight & Param. & Acc.  \\ \midrule
SimCLR  & ResNet50        & base & 32.16M & 69.24\% \\
        &   & ScoreCL & 33.54M & \textbf{72.26\%} \\ 
        & ResNet101      & base & 51.16M & 70.14\% \\
        &  & ScoreCL & 52.53M & \textbf{72.90\%} \\ \midrule
Simsiam & ResNet50       & base & 38.21M & 73.24\% \\
        &  & ScoreCL  & 39.59M & \textbf{74.18\%} \\ 
        & ResNet101      & base & 57.20M & 74.02\% \\
        &  & ScoreCL & 58.58M & \textbf{75.04\%} \\ \bottomrule
\end{tabular}
}
\caption{Comparison of the number of parameters and performance on ImageNet-100 dataset.}
\label{table:param}
\end{table}

\noindent\textbf{Ablation on False Positive Pair.} One may point out that false positive issues may arise, whereby an improperly augmented view is wrongly identified as a positive pair, for example, cropping the background without an object, yet treating it as a positive pair~\cite{purushwalkam2020demystifying, mo2021object, peng2022crafting}. 
We thus test whether our method can perform robustly with ContrastiveCrop~\cite{peng2022crafting}, which is proposed to solve the false positive problem. If our method assigns more penalties to the false positives, it may cause the ContrastiveCrop to fail. However, as shown in Table \ref{table:ab_fp}, there are further improvements when ScoreCL is applied even when ContrastiveCrop is used: it can boost performance due to its add-on property. 

\begin{wraptable}[7]{r}{0.5\textwidth}
    \centering
    \vspace{-5mm}
    \centering
    \begin{tabular}{c|ccc}
    \toprule
    Model   & Base  & Base+ & ScoreCL \\ \midrule
    SimCLR  & 60.11 & 60.91 & \textbf{62.34(+1.43)} \\
    Simsiam & 63.15 & 63.32 & \textbf{64.55(+1.23)} \\ \bottomrule
    \end{tabular}
    \caption{The performance of CLs with fair training time.}
    \label{table:cost}
\end{wraptable}

    

\noindent\textbf{Analysis on Training Costs. }The proposed score-guided CL has two training phase: score-matching function and CL. 
It is true that ours increases training cost, but it is minuscule since the score-matching function and CL are trained separately.
We show the number of parameters and performance in Table \ref{table:param}. 
Comparing ``\textit{ResNet50+score}" and ``\textit{ResNet101}" in each method, applying the proposed method achieves better performance even though there are fewer parameters of about 18M.
Note that the number of parameters in score mathcing function is about 1.8M and they are trained separately from CL.
Besides, Table \ref{table:cost} shows the performance of `base+' learned for the same amount of time as ScoreCL.

\begin{figure*}
    \centering
    \includegraphics[width=\columnwidth]{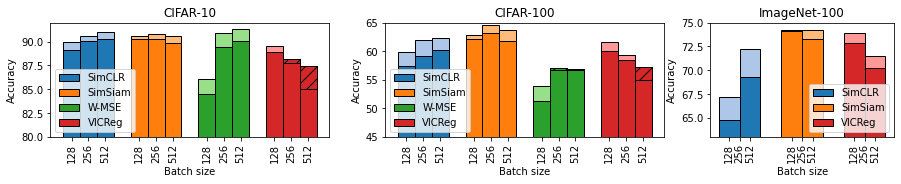}
    \caption{The ablation study on batch size for different datasets and CL models. Lighter colors represent an improved performance by applying ScoreCL. Instances of suboptimal performance are indicated by a diagonal slash.
    }
    \label{fig:bs_ablation}
\end{figure*}

\noindent\textbf{Ablation on Batch Size.} 
Depending on batch size, the performance of CL can be dramatically varied~\cite{chen2020simple}. 
Considering this, we investigate whether our method is effective for various batch sizes.
As illustrated in Fig. \ref{fig:bs_ablation}, even under any batch size condition except for VICReg on CIFAR-10 and CIFAR-100 datasets, it is shown that the performance is consistently enhanced when the score is applied to the existing CL. 
In the case of VICReg, it seems to be an unintended result because the variance regularizer can adversely affect training when images of the same class fit into one batch~\cite{bardes2021VICReg}.


\begin{figure}
    \centering
    \includegraphics[width=0.9\columnwidth]{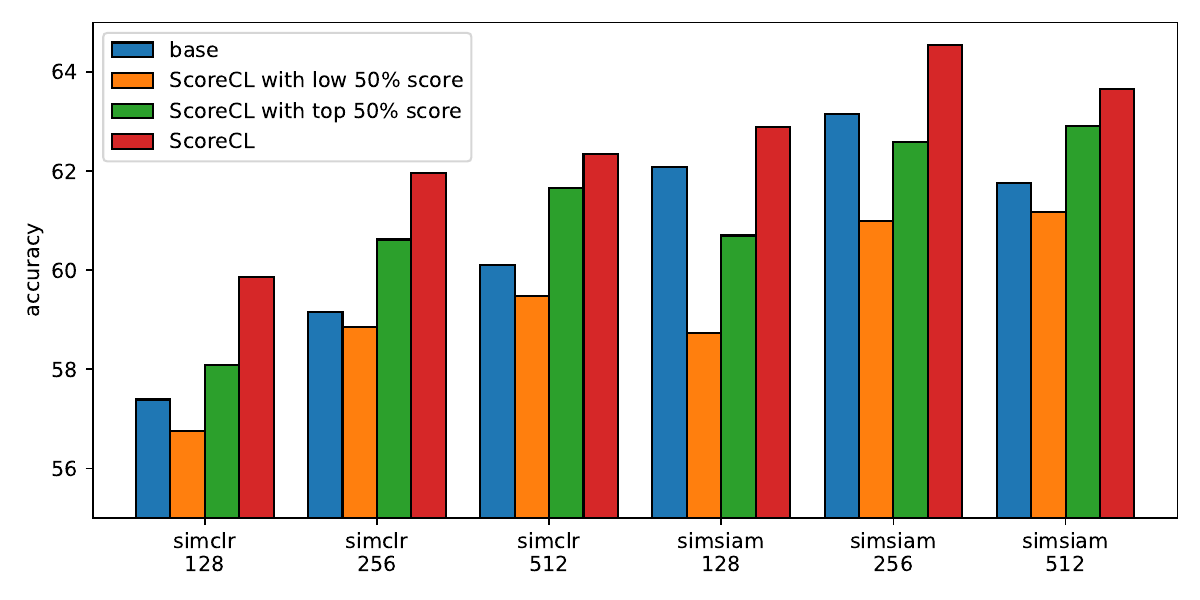}
    \caption{The ablation study on score thresholding with CIFAR-100 dataset. The integer numbers `\textit{128}', `\textit{256}', and `\textit{512}', indicate the batch size.
    }
    \label{fig:th_ablation}
\end{figure}

\noindent\textbf{Ablation on Sampling Strategy. }Unlike previous works that only focus on using view pairs having a large difference, our method can use view pairs with both large and small difference by penalizing the contrastive objective, allowing the CL to adaptively use a wide sample.
Here, we validate the effectiveness of this adaptive utilization scheme for a better understanding of our method.
To simulate that the only views with a large difference between variances are used in training CL, we sample the batch twice and use half of them by thresholding the scores with the median of score values of sampled batch data. 
The results illustrated in Fig.~\ref{fig:th_ablation} show that the score distance-agnostic method with a wide range of augmentation outperforms others only with a biased range.
\section{Conclusion}
In this paper, we propose a novel and simple approach for enhancing representation in CL with a score matching function. 
We tackle the issue of lacking contrastive objectives considering the view difference despite evidence supporting their efficacy in CL. 
Notably, it is the first work to analyze the property of the score matching function, linking them to augmentation intensity. 
Leveraging this insight,  we formulate ScoreCL that dynamically accommodates viewpoint diversity. 
Empirical evaluations underscore the consistent performance increase regardless of datasets, augmentation strategy, or CL models.
Addressing the false positive problem challenges inherent in CL augmentation, we extend our method to ConstrastCrop, yielding enhanced performance. 
Furthermore, our methods outperform baselines on downstream tasks for object classification and detection tasks.

Our proposed methods make CL model focus on the difference between the views to cover a wide range of view diversity. 
However, ScoreCL can fall into the risk of assigning the penalty considering only differences in augmentation regardless of the real class of images.
It implies that the representation of CL used in multiple downstream tasks can be collapsed, which can adversely affect other applications. To overcome this, class-specific methods, as well as augmentation scale-agnostic approach, should be studied as we do in Experiments section.
In order to overcome the limitation of showing only experimental evidence for the hypothesis about the score, we will verify the inferred property of score matching through theoretical analysis. 
In addition, by applying a universal augmentation technique and using an augmentation-agnostic proposed method, we want to provide a future direction that only needs to consider a contrastive objective among the significant components of CL.


\bibliography{sn-bibliography}


\begin{thebibliography}{60}
\ifx \bisbn   \undefined \def \bisbn  #1{ISBN #1}\fi
\ifx \binits  \undefined \def \binits#1{#1}\fi
\ifx \bauthor  \undefined \def \bauthor#1{#1}\fi
\ifx \batitle  \undefined \def \batitle#1{#1}\fi
\ifx \bjtitle  \undefined \def \bjtitle#1{#1}\fi
\ifx \bvolume  \undefined \def \bvolume#1{\textbf{#1}}\fi
\ifx \byear  \undefined \def \byear#1{#1}\fi
\ifx \bissue  \undefined \def \bissue#1{#1}\fi
\ifx \bfpage  \undefined \def \bfpage#1{#1}\fi
\ifx \blpage  \undefined \def \blpage #1{#1}\fi
\ifx \burl  \undefined \def \burl#1{\textsf{#1}}\fi
\ifx \doiurl  \undefined \def \doiurl#1{\url{https://doi.org/#1}}\fi
\ifx \betal  \undefined \def \betal{\textit{et al.}}\fi
\ifx \binstitute  \undefined \def \binstitute#1{#1}\fi
\ifx \binstitutionaled  \undefined \def \binstitutionaled#1{#1}\fi
\ifx \bctitle  \undefined \def \bctitle#1{#1}\fi
\ifx \beditor  \undefined \def \beditor#1{#1}\fi
\ifx \bpublisher  \undefined \def \bpublisher#1{#1}\fi
\ifx \bbtitle  \undefined \def \bbtitle#1{#1}\fi
\ifx \bedition  \undefined \def \bedition#1{#1}\fi
\ifx \bseriesno  \undefined \def \bseriesno#1{#1}\fi
\ifx \blocation  \undefined \def \blocation#1{#1}\fi
\ifx \bsertitle  \undefined \def \bsertitle#1{#1}\fi
\ifx \bsnm \undefined \def \bsnm#1{#1}\fi
\ifx \bsuffix \undefined \def \bsuffix#1{#1}\fi
\ifx \bparticle \undefined \def \bparticle#1{#1}\fi
\ifx \barticle \undefined \def \barticle#1{#1}\fi
\bibcommenthead
\ifx \bconfdate \undefined \def \bconfdate #1{#1}\fi
\ifx \botherref \undefined \def \botherref #1{#1}\fi
\ifx \url \undefined \def \url#1{\textsf{#1}}\fi
\ifx \bchapter \undefined \def \bchapter#1{#1}\fi
\ifx \bbook \undefined \def \bbook#1{#1}\fi
\ifx \bcomment \undefined \def \bcomment#1{#1}\fi
\ifx \oauthor \undefined \def \oauthor#1{#1}\fi
\ifx \citeauthoryear \undefined \def \citeauthoryear#1{#1}\fi
\ifx \endbibitem  \undefined \def \endbibitem {}\fi
\ifx \bconflocation  \undefined \def \bconflocation#1{#1}\fi
\ifx \arxivurl  \undefined \def \arxivurl#1{\textsf{#1}}\fi
\csname PreBibitemsHook\endcsname

\bibitem[\protect\citeauthoryear{Girshick et~al.}{2014}]{girshick2014rich}
\begin{bchapter}
\bauthor{\bsnm{Girshick}, \binits{R.}},
\bauthor{\bsnm{Donahue}, \binits{J.}},
\bauthor{\bsnm{Darrell}, \binits{T.}},
\bauthor{\bsnm{Malik}, \binits{J.}}:
\bctitle{Rich feature hierarchies for accurate object detection and semantic
  segmentation}.
In: \bbtitle{Proceedings of the IEEE Conference on Computer Vision and Pattern
  Recognition},
pp. \bfpage{580}--\blpage{587}
(\byear{2014})
\end{bchapter}
\endbibitem

\bibitem[\protect\citeauthoryear{Ren et~al.}{2015}]{ren2015faster}
\begin{botherref}
\oauthor{\bsnm{Ren}, \binits{S.}},
\oauthor{\bsnm{He}, \binits{K.}},
\oauthor{\bsnm{Girshick}, \binits{R.}},
\oauthor{\bsnm{Sun}, \binits{J.}}:
Faster r-cnn: Towards real-time object detection with region proposal networks.
Advances in neural information processing systems
\textbf{28}
(2015)
\end{botherref}
\endbibitem

\bibitem[\protect\citeauthoryear{Long et~al.}{2015}]{long2015fully}
\begin{bchapter}
\bauthor{\bsnm{Long}, \binits{J.}},
\bauthor{\bsnm{Shelhamer}, \binits{E.}},
\bauthor{\bsnm{Darrell}, \binits{T.}}:
\bctitle{Fully convolutional networks for semantic segmentation}.
In: \bbtitle{Proceedings of the IEEE Conference on Computer Vision and Pattern
  Recognition},
pp. \bfpage{3431}--\blpage{3440}
(\byear{2015})
\end{bchapter}
\endbibitem

\bibitem[\protect\citeauthoryear{He et~al.}{2017}]{he2017mask}
\begin{bchapter}
\bauthor{\bsnm{He}, \binits{K.}},
\bauthor{\bsnm{Gkioxari}, \binits{G.}},
\bauthor{\bsnm{Doll{\'a}r}, \binits{P.}},
\bauthor{\bsnm{Girshick}, \binits{R.}}:
\bctitle{Mask r-cnn}.
In: \bbtitle{Proceedings of the IEEE International Conference on Computer
  Vision},
pp. \bfpage{2961}--\blpage{2969}
(\byear{2017})
\end{bchapter}
\endbibitem

\bibitem[\protect\citeauthoryear{Krizhevsky
  et~al.}{2017}]{krizhevsky2017imagenet}
\begin{barticle}
\bauthor{\bsnm{Krizhevsky}, \binits{A.}},
\bauthor{\bsnm{Sutskever}, \binits{I.}},
\bauthor{\bsnm{Hinton}, \binits{G.E.}}:
\batitle{Imagenet classification with deep convolutional neural networks}.
\bjtitle{Communications of the ACM}
\bvolume{60}(\bissue{6}),
\bfpage{84}--\blpage{90}
(\byear{2017})
\end{barticle}
\endbibitem

\bibitem[\protect\citeauthoryear{Russakovsky
  et~al.}{2015}]{russakovsky2015imagenet}
\begin{barticle}
\bauthor{\bsnm{Russakovsky}, \binits{O.}},
\bauthor{\bsnm{Deng}, \binits{J.}},
\bauthor{\bsnm{Su}, \binits{H.}},
\bauthor{\bsnm{Krause}, \binits{J.}},
\bauthor{\bsnm{Satheesh}, \binits{S.}},
\bauthor{\bsnm{Ma}, \binits{S.}},
\bauthor{\bsnm{Huang}, \binits{Z.}},
\bauthor{\bsnm{Karpathy}, \binits{A.}},
\bauthor{\bsnm{Khosla}, \binits{A.}},
\bauthor{\bsnm{Bernstein}, \binits{M.}}, \betal:
\batitle{Imagenet large scale visual recognition challenge}.
\bjtitle{International journal of computer vision}
\bvolume{115}(\bissue{3}),
\bfpage{211}--\blpage{252}
(\byear{2015})
\end{barticle}
\endbibitem

\bibitem[\protect\citeauthoryear{Chen et~al.}{2020}]{chen2020simple}
\begin{bchapter}
\bauthor{\bsnm{Chen}, \binits{T.}},
\bauthor{\bsnm{Kornblith}, \binits{S.}},
\bauthor{\bsnm{Norouzi}, \binits{M.}},
\bauthor{\bsnm{Hinton}, \binits{G.}}:
\bctitle{A simple framework for contrastive learning of visual
  representations}.
In: \bbtitle{International Conference on Machine Learning},
pp. \bfpage{1597}--\blpage{1607}
(\byear{2020}).
\bcomment{PMLR}
\end{bchapter}
\endbibitem

\bibitem[\protect\citeauthoryear{Chen and He}{2021}]{chen2021exploring}
\begin{bchapter}
\bauthor{\bsnm{Chen}, \binits{X.}},
\bauthor{\bsnm{He}, \binits{K.}}:
\bctitle{Exploring simple siamese representation learning}.
In: \bbtitle{Proceedings of the IEEE/CVF Conference on Computer Vision and
  Pattern Recognition},
pp. \bfpage{15750}--\blpage{15758}
(\byear{2021})
\end{bchapter}
\endbibitem

\bibitem[\protect\citeauthoryear{Zbontar et~al.}{2021}]{zbontar2021barlow}
\begin{bchapter}
\bauthor{\bsnm{Zbontar}, \binits{J.}},
\bauthor{\bsnm{Jing}, \binits{L.}},
\bauthor{\bsnm{Misra}, \binits{I.}},
\bauthor{\bsnm{LeCun}, \binits{Y.}},
\bauthor{\bsnm{Deny}, \binits{S.}}:
\bctitle{Barlow twins: Self-supervised learning via redundancy reduction}.
In: \bbtitle{International Conference on Machine Learning},
pp. \bfpage{12310}--\blpage{12320}
(\byear{2021}).
\bcomment{PMLR}
\end{bchapter}
\endbibitem

\bibitem[\protect\citeauthoryear{Bardes et~al.}{2021}]{bardes2021VICReg}
\begin{botherref}
\oauthor{\bsnm{Bardes}, \binits{A.}},
\oauthor{\bsnm{Ponce}, \binits{J.}},
\oauthor{\bsnm{LeCun}, \binits{Y.}}:
Vicreg: Variance-invariance-covariance regularization for self-supervised
  learning.
arXiv preprint arXiv:2105.04906
(2021)
\end{botherref}
\endbibitem

\bibitem[\protect\citeauthoryear{Ermolov et~al.}{2021}]{ermolov2021whitening}
\begin{bchapter}
\bauthor{\bsnm{Ermolov}, \binits{A.}},
\bauthor{\bsnm{Siarohin}, \binits{A.}},
\bauthor{\bsnm{Sangineto}, \binits{E.}},
\bauthor{\bsnm{Sebe}, \binits{N.}}:
\bctitle{Whitening for self-supervised representation learning}.
In: \bbtitle{International Conference on Machine Learning},
pp. \bfpage{3015}--\blpage{3024}
(\byear{2021}).
\bcomment{PMLR}
\end{bchapter}
\endbibitem

\bibitem[\protect\citeauthoryear{He et~al.}{2020}]{he2020momentum}
\begin{bchapter}
\bauthor{\bsnm{He}, \binits{K.}},
\bauthor{\bsnm{Fan}, \binits{H.}},
\bauthor{\bsnm{Wu}, \binits{Y.}},
\bauthor{\bsnm{Xie}, \binits{S.}},
\bauthor{\bsnm{Girshick}, \binits{R.}}:
\bctitle{Momentum contrast for unsupervised visual representation learning}.
In: \bbtitle{Proceedings of the IEEE/CVF Conference on Computer Vision and
  Pattern Recognition},
pp. \bfpage{9729}--\blpage{9738}
(\byear{2020})
\end{bchapter}
\endbibitem

\bibitem[\protect\citeauthoryear{Chen et~al.}{2020}]{chen2020improved}
\begin{botherref}
\oauthor{\bsnm{Chen}, \binits{X.}},
\oauthor{\bsnm{Fan}, \binits{H.}},
\oauthor{\bsnm{Girshick}, \binits{R.}},
\oauthor{\bsnm{He}, \binits{K.}}:
Improved baselines with momentum contrastive learning.
arXiv preprint arXiv:2003.04297
(2020)
\end{botherref}
\endbibitem

\bibitem[\protect\citeauthoryear{Grill et~al.}{2020}]{grill2020bootstrap}
\begin{barticle}
\bauthor{\bsnm{Grill}, \binits{J.-B.}},
\bauthor{\bsnm{Strub}, \binits{F.}},
\bauthor{\bsnm{Altch{\'e}}, \binits{F.}},
\bauthor{\bsnm{Tallec}, \binits{C.}},
\bauthor{\bsnm{Richemond}, \binits{P.}},
\bauthor{\bsnm{Buchatskaya}, \binits{E.}},
\bauthor{\bsnm{Doersch}, \binits{C.}},
\bauthor{\bsnm{Avila~Pires}, \binits{B.}},
\bauthor{\bsnm{Guo}, \binits{Z.}},
\bauthor{\bsnm{Gheshlaghi~Azar}, \binits{M.}}, \betal:
\batitle{Bootstrap your own latent-a new approach to self-supervised learning}.
\bjtitle{Advances in neural information processing systems}
\bvolume{33},
\bfpage{21271}--\blpage{21284}
(\byear{2020})
\end{barticle}
\endbibitem

\bibitem[\protect\citeauthoryear{Caron et~al.}{2020}]{caron2020unsupervised}
\begin{barticle}
\bauthor{\bsnm{Caron}, \binits{M.}},
\bauthor{\bsnm{Misra}, \binits{I.}},
\bauthor{\bsnm{Mairal}, \binits{J.}},
\bauthor{\bsnm{Goyal}, \binits{P.}},
\bauthor{\bsnm{Bojanowski}, \binits{P.}},
\bauthor{\bsnm{Joulin}, \binits{A.}}:
\batitle{Unsupervised learning of visual features by contrasting cluster
  assignments}.
\bjtitle{Advances in Neural Information Processing Systems}
\bvolume{33},
\bfpage{9912}--\blpage{9924}
(\byear{2020})
\end{barticle}
\endbibitem

\bibitem[\protect\citeauthoryear{Li et~al.}{2022}]{li2022twin}
\begin{barticle}
\bauthor{\bsnm{Li}, \binits{Y.}},
\bauthor{\bsnm{Yang}, \binits{M.}},
\bauthor{\bsnm{Peng}, \binits{D.}},
\bauthor{\bsnm{Li}, \binits{T.}},
\bauthor{\bsnm{Huang}, \binits{J.}},
\bauthor{\bsnm{Peng}, \binits{X.}}:
\batitle{Twin contrastive learning for online clustering}.
\bjtitle{International Journal of Computer Vision}
\bvolume{130}(\bissue{9}),
\bfpage{2205}--\blpage{2221}
(\byear{2022})
\end{barticle}
\endbibitem

\bibitem[\protect\citeauthoryear{Wang and Qi}{2022}]{wang2022contrastive}
\begin{botherref}
\oauthor{\bsnm{Wang}, \binits{X.}},
\oauthor{\bsnm{Qi}, \binits{G.-J.}}:
Contrastive learning with stronger augmentations.
IEEE Transactions on Pattern Analysis and Machine Intelligence
(2022)
\end{botherref}
\endbibitem

\bibitem[\protect\citeauthoryear{Xie et~al.}{2022}]{xie2022delving}
\begin{barticle}
\bauthor{\bsnm{Xie}, \binits{J.}},
\bauthor{\bsnm{Zhan}, \binits{X.}},
\bauthor{\bsnm{Liu}, \binits{Z.}},
\bauthor{\bsnm{Ong}, \binits{Y.-S.}},
\bauthor{\bsnm{Loy}, \binits{C.C.}}:
\batitle{Delving into inter-image invariance for unsupervised visual
  representations}.
\bjtitle{International Journal of Computer Vision}
\bvolume{130}(\bissue{12}),
\bfpage{2994}--\blpage{3013}
(\byear{2022})
\end{barticle}
\endbibitem

\bibitem[\protect\citeauthoryear{Tian et~al.}{2020a}]{tian2020makes}
\begin{barticle}
\bauthor{\bsnm{Tian}, \binits{Y.}},
\bauthor{\bsnm{Sun}, \binits{C.}},
\bauthor{\bsnm{Poole}, \binits{B.}},
\bauthor{\bsnm{Krishnan}, \binits{D.}},
\bauthor{\bsnm{Schmid}, \binits{C.}},
\bauthor{\bsnm{Isola}, \binits{P.}}:
\batitle{What makes for good views for contrastive learning?}
\bjtitle{Advances in Neural Information Processing Systems}
\bvolume{33},
\bfpage{6827}--\blpage{6839}
(\byear{2020})
\end{barticle}
\endbibitem

\bibitem[\protect\citeauthoryear{Tian et~al.}{2020b}]{tian2020contrastive}
\begin{bchapter}
\bauthor{\bsnm{Tian}, \binits{Y.}},
\bauthor{\bsnm{Krishnan}, \binits{D.}},
\bauthor{\bsnm{Isola}, \binits{P.}}:
\bctitle{Contrastive multiview coding}.
In: \bbtitle{European Conference on Computer Vision},
pp. \bfpage{776}--\blpage{794}
(\byear{2020}).
\bcomment{Springer}
\end{bchapter}
\endbibitem

\bibitem[\protect\citeauthoryear{Wang et~al.}{2022}]{wang2022importance}
\begin{bchapter}
\bauthor{\bsnm{Wang}, \binits{X.}},
\bauthor{\bsnm{Fan}, \binits{H.}},
\bauthor{\bsnm{Tian}, \binits{Y.}},
\bauthor{\bsnm{Kihara}, \binits{D.}},
\bauthor{\bsnm{Chen}, \binits{X.}}:
\bctitle{On the importance of asymmetry for siamese representation learning}.
In: \bbtitle{Proceedings of the IEEE/CVF Conference on Computer Vision and
  Pattern Recognition},
pp. \bfpage{16570}--\blpage{16579}
(\byear{2022})
\end{bchapter}
\endbibitem

\bibitem[\protect\citeauthoryear{Peng et~al.}{2022}]{peng2022crafting}
\begin{bchapter}
\bauthor{\bsnm{Peng}, \binits{X.}},
\bauthor{\bsnm{Wang}, \binits{K.}},
\bauthor{\bsnm{Zhu}, \binits{Z.}},
\bauthor{\bsnm{Wang}, \binits{M.}},
\bauthor{\bsnm{You}, \binits{Y.}}:
\bctitle{Crafting better contrastive views for siamese representation
  learning}.
In: \bbtitle{Proceedings of the IEEE/CVF Conference on Computer Vision and
  Pattern Recognition},
pp. \bfpage{16031}--\blpage{16040}
(\byear{2022})
\end{bchapter}
\endbibitem

\bibitem[\protect\citeauthoryear{Hyv{\"a}rinen
  et~al.}{2009}]{hyvarinen2009estimation}
\begin{bchapter}
\bauthor{\bsnm{Hyv{\"a}rinen}, \binits{A.}},
\bauthor{\bsnm{Hurri}, \binits{J.}},
\bauthor{\bsnm{Hoyer}, \binits{P.O.}}:
\bctitle{Estimation of non-normalized statistical models}.
In: \bbtitle{Natural Image Statistics},
pp. \bfpage{419}--\blpage{426}.
\bpublisher{Springer}, \blocation{???}
(\byear{2009})
\end{bchapter}
\endbibitem

\bibitem[\protect\citeauthoryear{Hyv{\"a}rinen}{2008}]{hyvarinen2008optimal}
\begin{barticle}
\bauthor{\bsnm{Hyv{\"a}rinen}, \binits{A.}}:
\batitle{Optimal approximation of signal priors}.
\bjtitle{Neural Computation}
\bvolume{20}(\bissue{12}),
\bfpage{3087}--\blpage{3110}
(\byear{2008})
\end{barticle}
\endbibitem

\bibitem[\protect\citeauthoryear{Song et~al.}{2020}]{song2020sliced}
\begin{bchapter}
\bauthor{\bsnm{Song}, \binits{Y.}},
\bauthor{\bsnm{Garg}, \binits{S.}},
\bauthor{\bsnm{Shi}, \binits{J.}},
\bauthor{\bsnm{Ermon}, \binits{S.}}:
\bctitle{Sliced score matching: A scalable approach to density and score
  estimation}.
In: \bbtitle{Uncertainty in Artificial Intelligence},
pp. \bfpage{574}--\blpage{584}
(\byear{2020}).
\bcomment{PMLR}
\end{bchapter}
\endbibitem

\bibitem[\protect\citeauthoryear{Vincent}{2011}]{vincent2011connection}
\begin{barticle}
\bauthor{\bsnm{Vincent}, \binits{P.}}:
\batitle{A connection between score matching and denoising autoencoders}.
\bjtitle{Neural computation}
\bvolume{23}(\bissue{7}),
\bfpage{1661}--\blpage{1674}
(\byear{2011})
\end{barticle}
\endbibitem

\bibitem[\protect\citeauthoryear{Krizhevsky
  et~al.}{2009}]{krizhevsky2009learning}
\begin{botherref}
\oauthor{\bsnm{Krizhevsky}, \binits{A.}},
\oauthor{\bsnm{Hinton}, \binits{G.}}, et al.:
Learning multiple layers of features from tiny images
(2009)
\end{botherref}
\endbibitem

\bibitem[\protect\citeauthoryear{Deng et~al.}{2009}]{deng2009imagenet}
\begin{bchapter}
\bauthor{\bsnm{Deng}, \binits{J.}},
\bauthor{\bsnm{Dong}, \binits{W.}},
\bauthor{\bsnm{Socher}, \binits{R.}},
\bauthor{\bsnm{Li}, \binits{L.-J.}},
\bauthor{\bsnm{Li}, \binits{K.}},
\bauthor{\bsnm{Fei-Fei}, \binits{L.}}:
\bctitle{Imagenet: A large-scale hierarchical image database}.
In: \bbtitle{2009 IEEE Conference on Computer Vision and Pattern Recognition},
pp. \bfpage{248}--\blpage{255}
(\byear{2009}).
\bcomment{Ieee}
\end{bchapter}
\endbibitem

\bibitem[\protect\citeauthoryear{Oord et~al.}{2018}]{oord2018representation}
\begin{botherref}
\oauthor{\bsnm{Oord}, \binits{A.v.d.}},
\oauthor{\bsnm{Li}, \binits{Y.}},
\oauthor{\bsnm{Vinyals}, \binits{O.}}:
Representation learning with contrastive predictive coding.
arXiv preprint arXiv:1807.03748
(2018)
\end{botherref}
\endbibitem

\bibitem[\protect\citeauthoryear{Song and Ermon}{2020}]{song2020multicpc}
\begin{bchapter}
\bauthor{\bsnm{Song}, \binits{J.}},
\bauthor{\bsnm{Ermon}, \binits{S.}}:
\bctitle{Multi-label contrastive predictive coding}.
In: \beditor{\bsnm{Larochelle}, \binits{H.}},
\beditor{\bsnm{Ranzato}, \binits{M.}},
\beditor{\bsnm{Hadsell}, \binits{R.}},
\beditor{\bsnm{Balcan}, \binits{M.F.}},
\beditor{\bsnm{Lin}, \binits{H.}} (eds.)
\bbtitle{Advances in Neural Information Processing Systems},
vol. \bseriesno{33},
pp. \bfpage{8161}--\blpage{8173}.
\bpublisher{Curran Associates, Inc.}, \blocation{???}
(\byear{2020})
\end{bchapter}
\endbibitem

\bibitem[\protect\citeauthoryear{Huang et~al.}{2021}]{huang2021towards}
\begin{botherref}
\oauthor{\bsnm{Huang}, \binits{W.}},
\oauthor{\bsnm{Yi}, \binits{M.}},
\oauthor{\bsnm{Zhao}, \binits{X.}}:
Towards the generalization of contrastive self-supervised learning.
arXiv preprint arXiv:2111.00743
(2021)
\end{botherref}
\endbibitem

\bibitem[\protect\citeauthoryear{Zhang and Chen}{2021}]{zhang2021diffusion}
\begin{barticle}
\bauthor{\bsnm{Zhang}, \binits{Q.}},
\bauthor{\bsnm{Chen}, \binits{Y.}}:
\batitle{Diffusion normalizing flow}.
\bjtitle{Advances in Neural Information Processing Systems}
\bvolume{34},
\bfpage{16280}--\blpage{16291}
(\byear{2021})
\end{barticle}
\endbibitem

\bibitem[\protect\citeauthoryear{Gong and Li}{2021}]{gong2021interpreting}
\begin{botherref}
\oauthor{\bsnm{Gong}, \binits{W.}},
\oauthor{\bsnm{Li}, \binits{Y.}}:
Interpreting diffusion score matching using normalizing flow.
arXiv preprint arXiv:2107.10072
(2021)
\end{botherref}
\endbibitem

\bibitem[\protect\citeauthoryear{Song and Ermon}{2019}]{song2019generative}
\begin{botherref}
\oauthor{\bsnm{Song}, \binits{Y.}},
\oauthor{\bsnm{Ermon}, \binits{S.}}:
Generative modeling by estimating gradients of the data distribution.
Advances in Neural Information Processing Systems
\textbf{32}
(2019)
\end{botherref}
\endbibitem

\bibitem[\protect\citeauthoryear{Song et~al.}{2020}]{song2020score}
\begin{botherref}
\oauthor{\bsnm{Song}, \binits{Y.}},
\oauthor{\bsnm{Sohl-Dickstein}, \binits{J.}},
\oauthor{\bsnm{Kingma}, \binits{D.P.}},
\oauthor{\bsnm{Kumar}, \binits{A.}},
\oauthor{\bsnm{Ermon}, \binits{S.}},
\oauthor{\bsnm{Poole}, \binits{B.}}:
Score-based generative modeling through stochastic differential equations.
arXiv preprint arXiv:2011.13456
(2020)
\end{botherref}
\endbibitem

\bibitem[\protect\citeauthoryear{Khosla et~al.}{2020}]{khosla2020supervised}
\begin{barticle}
\bauthor{\bsnm{Khosla}, \binits{P.}},
\bauthor{\bsnm{Teterwak}, \binits{P.}},
\bauthor{\bsnm{Wang}, \binits{C.}},
\bauthor{\bsnm{Sarna}, \binits{A.}},
\bauthor{\bsnm{Tian}, \binits{Y.}},
\bauthor{\bsnm{Isola}, \binits{P.}},
\bauthor{\bsnm{Maschinot}, \binits{A.}},
\bauthor{\bsnm{Liu}, \binits{C.}},
\bauthor{\bsnm{Krishnan}, \binits{D.}}:
\batitle{Supervised contrastive learning}.
\bjtitle{Advances in Neural Information Processing Systems}
\bvolume{33},
\bfpage{18661}--\blpage{18673}
(\byear{2020})
\end{barticle}
\endbibitem

\bibitem[\protect\citeauthoryear{Henaff}{2020}]{henaff2020data}
\begin{bchapter}
\bauthor{\bsnm{Henaff}, \binits{O.}}:
\bctitle{Data-efficient image recognition with contrastive predictive coding}.
In: \bbtitle{International Conference on Machine Learning},
pp. \bfpage{4182}--\blpage{4192}
(\byear{2020}).
\bcomment{PMLR}
\end{bchapter}
\endbibitem

\bibitem[\protect\citeauthoryear{Mahmood et~al.}{2020}]{mahmood2020multiscale}
\begin{botherref}
\oauthor{\bsnm{Mahmood}, \binits{A.}},
\oauthor{\bsnm{Oliva}, \binits{J.}},
\oauthor{\bsnm{Styner}, \binits{M.}}:
Multiscale score matching for out-of-distribution detection.
arXiv preprint arXiv:2010.13132
(2020)
\end{botherref}
\endbibitem

\bibitem[\protect\citeauthoryear{Lee and Shin}{2022}]{lee2022r}
\begin{botherref}
\oauthor{\bsnm{Lee}, \binits{K.}},
\oauthor{\bsnm{Shin}, \binits{J.}}:
R$\backslash$'enyicl: Contrastive representation learning with skew
  r$\backslash$'enyi divergence.
arXiv preprint arXiv:2208.06270
(2022)
\end{botherref}
\endbibitem

\bibitem[\protect\citeauthoryear{Robinson
  et~al.}{2020}]{robinson2020contrastive}
\begin{botherref}
\oauthor{\bsnm{Robinson}, \binits{J.}},
\oauthor{\bsnm{Chuang}, \binits{C.-Y.}},
\oauthor{\bsnm{Sra}, \binits{S.}},
\oauthor{\bsnm{Jegelka}, \binits{S.}}:
Contrastive learning with hard negative samples.
arXiv preprint arXiv:2010.04592
(2020)
\end{botherref}
\endbibitem

\bibitem[\protect\citeauthoryear{Cubuk et~al.}{2020}]{cubuk2020randaugment}
\begin{bchapter}
\bauthor{\bsnm{Cubuk}, \binits{E.D.}},
\bauthor{\bsnm{Zoph}, \binits{B.}},
\bauthor{\bsnm{Shlens}, \binits{J.}},
\bauthor{\bsnm{Le}, \binits{Q.V.}}:
\bctitle{Randaugment: Practical automated data augmentation with a reduced
  search space}.
In: \bbtitle{Proceedings of the IEEE/CVF Conference on Computer Vision and
  Pattern Recognition Workshops},
pp. \bfpage{702}--\blpage{703}
(\byear{2020})
\end{bchapter}
\endbibitem

\bibitem[\protect\citeauthoryear{Kadkhodaie and
  Simoncelli}{2021}]{kadkhodaie2021stochastic}
\begin{barticle}
\bauthor{\bsnm{Kadkhodaie}, \binits{Z.}},
\bauthor{\bsnm{Simoncelli}, \binits{E.}}:
\batitle{Stochastic solutions for linear inverse problems using the prior
  implicit in a denoiser}.
\bjtitle{Advances in Neural Information Processing Systems}
\bvolume{34},
\bfpage{13242}--\blpage{13254}
(\byear{2021})
\end{barticle}
\endbibitem

\bibitem[\protect\citeauthoryear{Bansal et~al.}{2022}]{bansal2022cold}
\begin{botherref}
\oauthor{\bsnm{Bansal}, \binits{A.}},
\oauthor{\bsnm{Borgnia}, \binits{E.}},
\oauthor{\bsnm{Chu}, \binits{H.-M.}},
\oauthor{\bsnm{Li}, \binits{J.S.}},
\oauthor{\bsnm{Kazemi}, \binits{H.}},
\oauthor{\bsnm{Huang}, \binits{F.}},
\oauthor{\bsnm{Goldblum}, \binits{M.}},
\oauthor{\bsnm{Geiping}, \binits{J.}},
\oauthor{\bsnm{Goldstein}, \binits{T.}}:
Cold diffusion: Inverting arbitrary image transforms without noise.
arXiv preprint arXiv:2208.09392
(2022)
\end{botherref}
\endbibitem

\bibitem[\protect\citeauthoryear{Sohl-Dickstein et~al.}{2015}]{sohl2015deep}
\begin{bchapter}
\bauthor{\bsnm{Sohl-Dickstein}, \binits{J.}},
\bauthor{\bsnm{Weiss}, \binits{E.}},
\bauthor{\bsnm{Maheswaranathan}, \binits{N.}},
\bauthor{\bsnm{Ganguli}, \binits{S.}}:
\bctitle{Deep unsupervised learning using nonequilibrium thermodynamics}.
In: \bbtitle{International Conference on Machine Learning},
pp. \bfpage{2256}--\blpage{2265}
(\byear{2015}).
\bcomment{PMLR}
\end{bchapter}
\endbibitem

\bibitem[\protect\citeauthoryear{Sutskever
  et~al.}{2013}]{sutskever2013importance}
\begin{bchapter}
\bauthor{\bsnm{Sutskever}, \binits{I.}},
\bauthor{\bsnm{Martens}, \binits{J.}},
\bauthor{\bsnm{Dahl}, \binits{G.}},
\bauthor{\bsnm{Hinton}, \binits{G.}}:
\bctitle{On the importance of initialization and momentum in deep learning}.
In: \bbtitle{International Conference on Machine Learning},
pp. \bfpage{1139}--\blpage{1147}
(\byear{2013}).
\bcomment{PMLR}
\end{bchapter}
\endbibitem

\bibitem[\protect\citeauthoryear{Kingma and Ba}{2014}]{kingma2014adam}
\begin{botherref}
\oauthor{\bsnm{Kingma}, \binits{D.P.}},
\oauthor{\bsnm{Ba}, \binits{J.}}:
Adam: A method for stochastic optimization.
arXiv preprint arXiv:1412.6980
(2014)
\end{botherref}
\endbibitem

\bibitem[\protect\citeauthoryear{You et~al.}{2019}]{you2019large}
\begin{botherref}
\oauthor{\bsnm{You}, \binits{Y.}},
\oauthor{\bsnm{Li}, \binits{J.}},
\oauthor{\bsnm{Reddi}, \binits{S.}},
\oauthor{\bsnm{Hseu}, \binits{J.}},
\oauthor{\bsnm{Kumar}, \binits{S.}},
\oauthor{\bsnm{Bhojanapalli}, \binits{S.}},
\oauthor{\bsnm{Song}, \binits{X.}},
\oauthor{\bsnm{Demmel}, \binits{J.}},
\oauthor{\bsnm{Keutzer}, \binits{K.}},
\oauthor{\bsnm{Hsieh}, \binits{C.-J.}}:
Large batch optimization for deep learning: Training bert in 76 minutes.
arXiv preprint arXiv:1904.00962
(2019)
\end{botherref}
\endbibitem

\bibitem[\protect\citeauthoryear{Wu et~al.}{2019}]{detectron2}
\begin{botherref}
\oauthor{\bsnm{Wu}, \binits{Y.}},
\oauthor{\bsnm{Kirillov}, \binits{A.}},
\oauthor{\bsnm{Massa}, \binits{F.}},
\oauthor{\bsnm{Lo}, \binits{W.-Y.}},
\oauthor{\bsnm{Girshick}, \binits{R.}}:
Detectron2.
\url{https://github.com/facebookresearch/detectron2}
(2019)
\end{botherref}
\endbibitem

\bibitem[\protect\citeauthoryear{Dosovitskiy et~al.}{2021}]{dosovitskiy2021an}
\begin{bchapter}
\bauthor{\bsnm{Dosovitskiy}, \binits{A.}},
\bauthor{\bsnm{Beyer}, \binits{L.}},
\bauthor{\bsnm{Kolesnikov}, \binits{A.}},
\bauthor{\bsnm{Weissenborn}, \binits{D.}},
\bauthor{\bsnm{Zhai}, \binits{X.}},
\bauthor{\bsnm{Unterthiner}, \binits{T.}},
\bauthor{\bsnm{Dehghani}, \binits{M.}},
\bauthor{\bsnm{Minderer}, \binits{M.}},
\bauthor{\bsnm{Heigold}, \binits{G.}},
\bauthor{\bsnm{Gelly}, \binits{S.}},
\bauthor{\bsnm{Uszkoreit}, \binits{J.}},
\bauthor{\bsnm{Houlsby}, \binits{N.}}:
\bctitle{An image is worth 16x16 words: Transformers for image recognition at
  scale}.
In: \bbtitle{International Conference on Learning Representations}
(\byear{2021}).
\burl{https://openreview.net/forum?id=YicbFdNTTy}
\end{bchapter}
\endbibitem

\bibitem[\protect\citeauthoryear{Coates et~al.}{2011}]{coates2011analysis}
\begin{bchapter}
\bauthor{\bsnm{Coates}, \binits{A.}},
\bauthor{\bsnm{Ng}, \binits{A.}},
\bauthor{\bsnm{Lee}, \binits{H.}}:
\bctitle{An analysis of single-layer networks in unsupervised feature
  learning}.
In: \bbtitle{Proceedings of the Fourteenth International Conference on
  Artificial Intelligence and Statistics},
pp. \bfpage{215}--\blpage{223}
(\byear{2011}).
\bcomment{JMLR Workshop and Conference Proceedings}
\end{bchapter}
\endbibitem

\bibitem[\protect\citeauthoryear{Bossard et~al.}{2014}]{bossard2014food}
\begin{bchapter}
\bauthor{\bsnm{Bossard}, \binits{L.}},
\bauthor{\bsnm{Guillaumin}, \binits{M.}},
\bauthor{\bsnm{Gool}, \binits{L.V.}}:
\bctitle{Food-101--mining discriminative components with random forests}.
In: \bbtitle{European Conference on Computer Vision},
pp. \bfpage{446}--\blpage{461}
(\byear{2014}).
\bcomment{Springer}
\end{bchapter}
\endbibitem

\bibitem[\protect\citeauthoryear{Nilsback and
  Zisserman}{2008}]{nilsback2008automated}
\begin{bchapter}
\bauthor{\bsnm{Nilsback}, \binits{M.-E.}},
\bauthor{\bsnm{Zisserman}, \binits{A.}}:
\bctitle{Automated flower classification over a large number of classes}.
In: \bbtitle{2008 Sixth Indian Conference on Computer Vision, Graphics \& Image
  Processing},
pp. \bfpage{722}--\blpage{729}
(\byear{2008}).
\bcomment{IEEE}
\end{bchapter}
\endbibitem

\bibitem[\protect\citeauthoryear{Krause et~al.}{2013}]{krause20133d}
\begin{bchapter}
\bauthor{\bsnm{Krause}, \binits{J.}},
\bauthor{\bsnm{Stark}, \binits{M.}},
\bauthor{\bsnm{Deng}, \binits{J.}},
\bauthor{\bsnm{Fei-Fei}, \binits{L.}}:
\bctitle{3d object representations for fine-grained categorization}.
In: \bbtitle{Proceedings of the IEEE International Conference on Computer
  Vision Workshops},
pp. \bfpage{554}--\blpage{561}
(\byear{2013})
\end{bchapter}
\endbibitem

\bibitem[\protect\citeauthoryear{Maji et~al.}{2013}]{maji13fine-grained}
\begin{botherref}
\oauthor{\bsnm{Maji}, \binits{S.}},
\oauthor{\bsnm{Kannala}, \binits{J.}},
\oauthor{\bsnm{Rahtu}, \binits{E.}},
\oauthor{\bsnm{Blaschko}, \binits{M.}},
\oauthor{\bsnm{Vedaldi}, \binits{A.}}:
Fine-grained visual classification of aircraft.
Technical report
(2013)
\end{botherref}
\endbibitem

\bibitem[\protect\citeauthoryear{Cimpoi et~al.}{2014}]{cimpoi14describing}
\begin{bchapter}
\bauthor{\bsnm{Cimpoi}, \binits{M.}},
\bauthor{\bsnm{Maji}, \binits{S.}},
\bauthor{\bsnm{Kokkinos}, \binits{I.}},
\bauthor{\bsnm{Mohamed}, \binits{S.}},
\bauthor{},
\bauthor{\bsnm{Vedaldi}, \binits{A.}}:
\bctitle{Describing textures in the wild}.
In: \bbtitle{Proceedings of the {IEEE} Conf. on Computer Vision and Pattern
  Recognition ({CVPR})}
(\byear{2014})
\end{bchapter}
\endbibitem

\bibitem[\protect\citeauthoryear{Everingham
  et~al.}{2010}]{everingham2010pascal}
\begin{barticle}
\bauthor{\bsnm{Everingham}, \binits{M.}},
\bauthor{\bsnm{Van~Gool}, \binits{L.}},
\bauthor{\bsnm{Williams}, \binits{C.K.}},
\bauthor{\bsnm{Winn}, \binits{J.}},
\bauthor{\bsnm{Zisserman}, \binits{A.}}:
\batitle{The pascal visual object classes (voc) challenge}.
\bjtitle{International journal of computer vision}
\bvolume{88}(\bissue{2}),
\bfpage{303}--\blpage{338}
(\byear{2010})
\end{barticle}
\endbibitem

\bibitem[\protect\citeauthoryear{Lin et~al.}{2014}]{lin2014microsoft}
\begin{bchapter}
\bauthor{\bsnm{Lin}, \binits{T.-Y.}},
\bauthor{\bsnm{Maire}, \binits{M.}},
\bauthor{\bsnm{Belongie}, \binits{S.}},
\bauthor{\bsnm{Hays}, \binits{J.}},
\bauthor{\bsnm{Perona}, \binits{P.}},
\bauthor{\bsnm{Ramanan}, \binits{D.}},
\bauthor{\bsnm{Doll{\'a}r}, \binits{P.}},
\bauthor{\bsnm{Zitnick}, \binits{C.L.}}:
\bctitle{Microsoft coco: Common objects in context}.
In: \bbtitle{European Conference on Computer Vision},
pp. \bfpage{740}--\blpage{755}
(\byear{2014}).
\bcomment{Springer}
\end{bchapter}
\endbibitem

\bibitem[\protect\citeauthoryear{Zhang et~al.}{2018}]{zhang2018unreasonable}
\begin{bchapter}
\bauthor{\bsnm{Zhang}, \binits{R.}},
\bauthor{\bsnm{Isola}, \binits{P.}},
\bauthor{\bsnm{Efros}, \binits{A.A.}},
\bauthor{\bsnm{Shechtman}, \binits{E.}},
\bauthor{\bsnm{Wang}, \binits{O.}}:
\bctitle{The unreasonable effectiveness of deep features as a perceptual
  metric}.
In: \bbtitle{Proceedings of the IEEE Conference on Computer Vision and Pattern
  Recognition},
pp. \bfpage{586}--\blpage{595}
(\byear{2018})
\end{bchapter}
\endbibitem

\bibitem[\protect\citeauthoryear{Purushwalkam and
  Gupta}{2020}]{purushwalkam2020demystifying}
\begin{barticle}
\bauthor{\bsnm{Purushwalkam}, \binits{S.}},
\bauthor{\bsnm{Gupta}, \binits{A.}}:
\batitle{Demystifying contrastive self-supervised learning: Invariances,
  augmentations and dataset biases}.
\bjtitle{Advances in Neural Information Processing Systems}
\bvolume{33},
\bfpage{3407}--\blpage{3418}
(\byear{2020})
\end{barticle}
\endbibitem

\bibitem[\protect\citeauthoryear{Mo et~al.}{2021}]{mo2021object}
\begin{barticle}
\bauthor{\bsnm{Mo}, \binits{S.}},
\bauthor{\bsnm{Kang}, \binits{H.}},
\bauthor{\bsnm{Sohn}, \binits{K.}},
\bauthor{\bsnm{Li}, \binits{C.-L.}},
\bauthor{\bsnm{Shin}, \binits{J.}}:
\batitle{Object-aware contrastive learning for debiased scene representation}.
\bjtitle{Advances in Neural Information Processing Systems}
\bvolume{34},
\bfpage{12251}--\blpage{12264}
(\byear{2021})
\end{barticle}
\endbibitem

\end{thebibliography}

\end{document}